\documentclass[conference, twocolumn]{IEEEtran}
\usepackage{listings}
\usepackage{graphicx}
\usepackage{subcaption}
\usepackage[table,xcdraw]{xcolor}
\usepackage{amsmath}
\usepackage{pifont}
\usepackage{amssymb}
\definecolor{mydarkblue}{rgb}{0,0.08,0.65}
\usepackage[colorlinks=true,
    linkcolor=mydarkblue,
    citecolor=mydarkblue,
    filecolor=mydarkblue,
    urlcolor=mydarkblue]{hyperref} 
\usepackage{cleveref}
\usepackage{soul}
\usepackage[shortlabels]{enumitem}
\usepackage{url}
\usepackage{color}
\usepackage{tikz}
\usepackage{balance}
\usepackage{multirow}
\usepackage{comment}
\usepackage{wrapfig,lipsum,booktabs}
\usepackage[frozencache,cachedir=.]{minted}

\usepackage[symbol]{footmisc}
\usepackage{tablefootnote}
\usepackage{tabularx}
\usepackage{placeins}
\usepackage{algorithm}
\usepackage{algpseudocode}

\usepackage{cuted}
\usepackage{float}
\usepackage{caption}

\usepackage{multicol}
\usepackage{dblfloatfix}
\usepackage{natbib}

\definecolor{codegreen}{rgb}{0,0.6,0}
\definecolor{codegray}{rgb}{0.5,0.5,0.5}
\definecolor{codepurple}{rgb}{0.58,0,0.82}
\definecolor{backcolour}{rgb}{0.95,0.95,0.92}

\makeatletter
\def\blfootnote{\xdef\@thefnmark{}\@footnotetext}
\makeatother

\lstdefinestyle{mystyle}{
  backgroundcolor=\color{backcolour},   commentstyle=\color{codegreen},
  keywordstyle=\color{magenta},
  numberstyle=\tiny\color{codegray},
  stringstyle=\color{codepurple},
  basicstyle=\ttfamily\footnotesize,
  breakatwhitespace=false,         
  breaklines=true,                 
  captionpos=b,                    
  keepspaces=true,                 
  numbers=left,                    
  numbersep=5pt,                  
  showspaces=false,                
  showstringspaces=false,
  showtabs=false,                  
  tabsize=2,
}

\AtBeginDocument{%
  \providecommand\BibTeX{{%
    \normalfont B\kern-0.5em{\scshape i\kern-0.25em b}\kern-0.8em\TeX}}}

\IEEEoverridecommandlockouts
\begin{document}

\title{Folding Tensor and Sequence Parallelism for Memory-Efficient Transformer Training \& Inference}

\newcommand{\corr}{\textsuperscript{*}}

\author{
\IEEEauthorblockN{Vasu Shyam, Anna Golubeva, Quentin Anthony}\\[0.6em]
\IEEEauthorblockA{\textbf{Zyphra}}
}

\author{
Vasu Shyam $\quad$
Anna Golubeva $\quad$
Quentin Anthony \\[0.6em]
{\textbf{Zyphra}}
}

\maketitle

\blfootnote{\textsuperscript{*}Corresponding authors: \texttt{vasu@zyphra.com}, \texttt{anna@zyphra.com}, \texttt{quentin@zyphra.com}}

\setcounter{page}{1}

\begin{abstract}
We present tensor and sequence parallelism (TSP), a parallel execution strategy that folds tensor parallelism and sequence parallelism onto a single device axis. In conventional multi-dimensional parallelism layouts, tensor parallelism (TP) shards model weights while sequence parallelism (SP) shards tokens, reducing per-device parameter or activation memory, respectively. Traditionally, each scheme is assigned its own mesh dimension. TSP instead assigns each rank both a weight shard and a sequence shard, reducing both parameter and activation memory along the same device axis. We implement this design with two runtime schedules. For attention, ranks iterate over broadcast parameter shards and reconstruct context through a sequence-wise key/value exchange. For gated MLPs, weight shards circulate in a ring while partial outputs accumulate locally. By sharding both weights and activations across the same devices, TSP trades additional communication volume for reduced memory overhead. We provide a theoretical communication and memory analysis, describe our implementation of TSP attention and gated MLP blocks, and benchmark TSP against TP, SP, and TP+SP. These results position TSP as a hardware-aware alternative for long-context and memory-constrained model training, and as a viable axis of parallelism in concert with existing parallelism schemes such as pipeline and expert parallelism for dense and mixture-of-expert models.
\end{abstract}

\section{Introduction}

Multidimensional parallelism is a key paradigm for scaling AI training and inference across modern HPC clusters. For large language models, the principal forms of parallelism are tensor (TP), pipeline (PP), expert (EP), data (DP), and sequence parallelism (SP). We focus on tensor and sequence parallelism and present a method for folding them both onto a single device-mesh axis. In tensor parallelism, the weight matrices of each layer are partitioned across devices. Each rank stores and applies a shard of every linear projection, and the partial results are reconciled through collective communication, typically an all-reduce or a reduce-scatter/all-gather pair. This reduces per-device parameter memory (including gradients and optimizer states) in proportion to the TP degree. Sequence parallelism instead partitions the input along the sequence dimension, so each rank holds and attends over only a fraction of the sequence. This reduces per-device activation memory and distributes the quadratic-in-sequence-length attention cost, but leaves model parameters fully replicated.

\begin{figure*}[t]
    \centering
    \includegraphics[width=\textwidth]{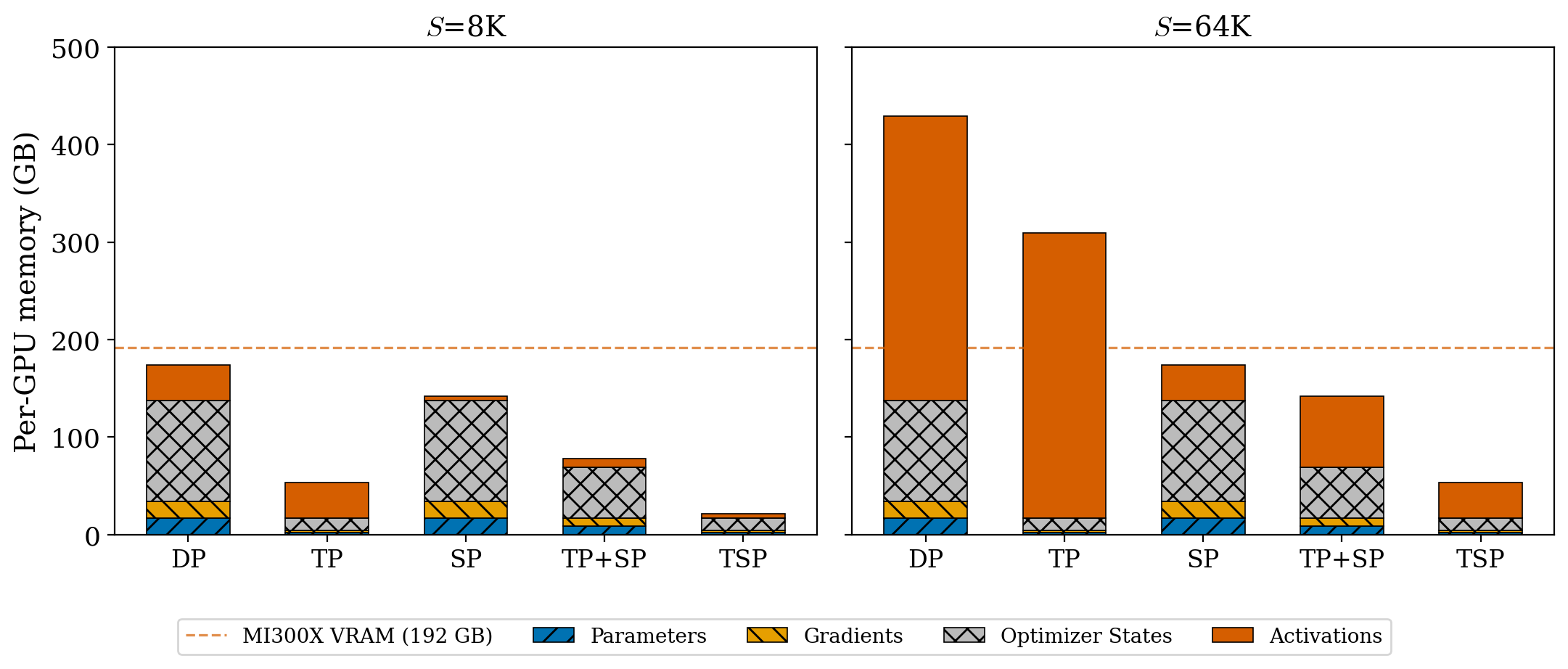}
    \caption{Per-GPU theoretical memory breakdown by parallelism strategy for a 7B-parameter dense transformer model with selective recomputation ($B=1$). All schemes are evaluated on a single 8-GPU node ($D=8$): TP uses $T=8$, SP uses $\Sigma=8$, TSP uses $D_{\mathrm{TSP}}=8$, and TP+SP uses a balanced mesh with $T=2$, $\Sigma=4$. At short context~(left, $S{=}8192$), weight-proportional memory (parameters, gradients, optimizer states) dominates and TP or TSP provide the largest reduction. At long context~(right, $S{=}65536$), activation memory becomes the bottleneck. The dotted line represents the 192~GB VRAM of the MI300X GPU; bars crossing this line will OOM.}
    \label{fig:mem-bars}
\end{figure*}

In the standard multidimensional-parallelism framework, a logical mesh is formed from the full set of devices and each parallelism strategy is assigned its own orthogonal axis. Because the total device count is fixed ($N = T \cdot \Sigma \cdot N_{\text{DP}} \cdot N_{\text{PP}}$, where $T$ and $\Sigma$ are the tensor- and sequence-parallel degrees respectively) every rank allocated to one axis reduces the ranks available for the others. In particular, using separate axes for TP and SP allocates $T \cdot \Sigma$ ranks to model parallelism, leaving correspondingly fewer replicas for data parallelism.

Here we show that the TP and SP axes can be collapsed into a single axis of size $N_{\text{TSP}}$, with each rank holding both a shard of the sequence and a shard of the model weights. This yields the per-device memory benefits of both forms of parallelism, the benefits being reduced parameter memory from weight sharding and reduced activation memory from sequence sharding, while freeing the ranks that would otherwise form the second orthogonal axis to serve as additional data-parallel replicas.

Concretely, for multi-head attention, each iteration of the TSP loop broadcasts one rank's weight shards (the $W_Q, W_K, W_V, W_O$ projections, packed into a single buffer) to all peers. Every rank then computes local projections on its sequence shard and participates in an all-gather of the keys and values to reconstruct the full sequence context. A zigzag partitioning scheme balances the causal-attention workload across ranks. For the MLP, the analogy between $\sigma(X \cdot W_1^\top)\, W_2$ and $\mathrm{softmax}(Q \cdot K^\top)\, V$ motivates a ring-communication schedule. Weight shards circulate via point-to-point send/recv while each rank accumulates partial outputs locally, eliminating the all-reduce that standard TP requires.

A further benefit is topological: Because $N_{\text{TSP}} \le N_{\text{TP}} \cdot N_{\text{SP}}$, the combined group can be mapped entirely within the high-bandwidth intra-node domain (NVLink or AMD Infinity Fabric). The conventional two-axis layout requires $N_{\text{TP}} \cdot N_{\text{SP}}$ ranks for the same degree of tensor and sequence sharding, which in many configurations spills onto lower-bandwidth inter-node interconnects (InfiniBand, Ethernet) and under-utilizes the available bisection bandwidth.

\section{Background}
\label{sec:background}

Distributed model training relies on combining multiple forms of parallelism across a fixed set of accelerator devices. Each parallelism scheme incurs its own trade-offs, and the exact mesh dimensions must be chosen based on the combination of model topology and cluster hardware. We focus on two forms of parallelism: tensor parallelism (TP), which shards weights across ranks, and sequence parallelism (SP), which shards the token sequence across ranks.

At a high level, a layer's communication pattern is determined by which tensors (weights, activations, gradients, and optimizer states) are replicated and which are sharded. If \textit{weights} are sharded, each rank computes only a portion of the weights' results of the full sequence, and all ranks' partial results must be shared via collective communication operations. If the \textit{sequence} is sharded, each rank stores only a subset of tokens, and any sequence-dependent operations such as attention will require collective operations to perform the full operation across the full sequence. This perspective makes TP and SP complementary: TP shards across weights, while SP shards across the sequence.

\begin{figure*}[hbtp]
    \centering
    \includegraphics[width=0.90\textwidth]{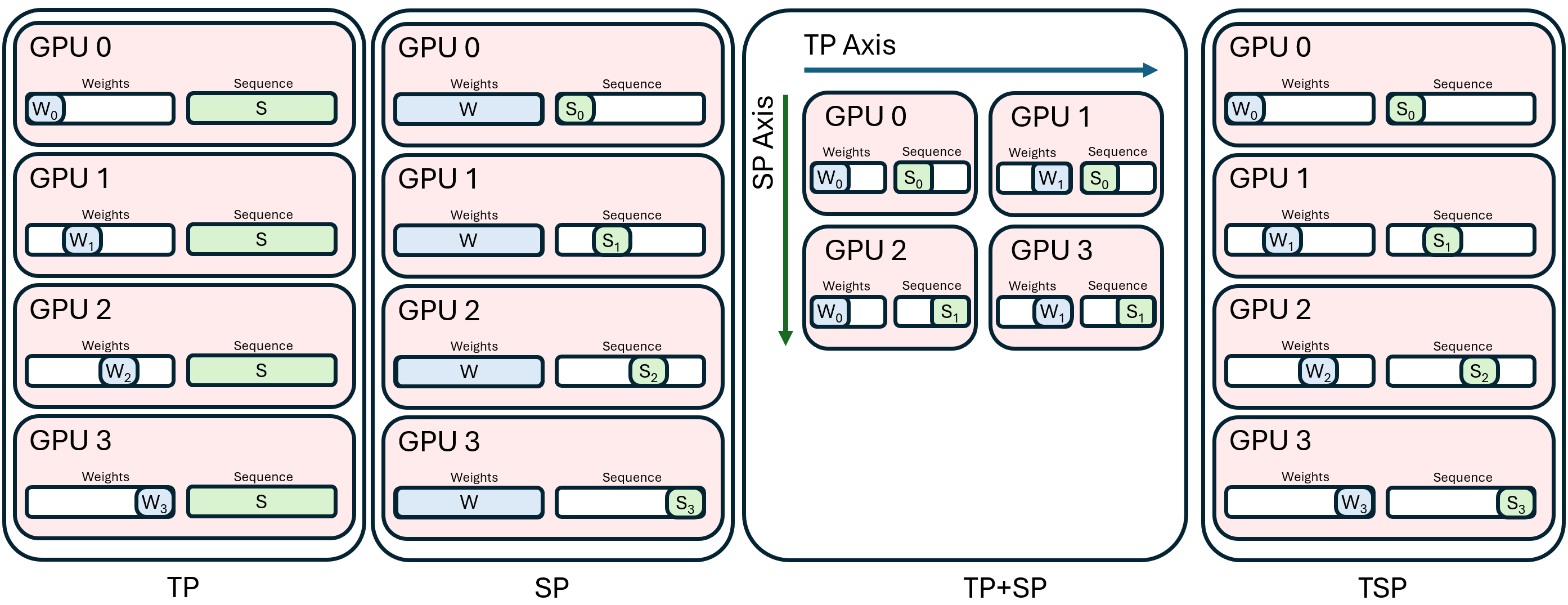}
    \caption{Weights and sequence sharding under TP, SP, TP+SP, and TSP. Note that TP+SP uses two orthogonal axes to split the parallel groups while TSP lets both weights and activations be sharded simultaneously along the same devices.}
    \label{fig:tsp-background}
\end{figure*}

\subsection{Tensor Parallelism}
\label{sec:background-tp}

Tensor parallelism partitions the parameters of a layer across multiple devices while keeping the inputs replicated. In a Transformer block, this is typically implemented by sharding the output dimension of projection matrices such as $W_Q$, $W_K$, $W_V$, and the up and gate projections of the MLP, followed by a row-parallel output projection whose partial results are summed across ranks. For a linear map $Y = XW^\top$, each rank stores only a shard of $W$, computes its local contribution, and then participates in an all-reduce to form the full output.

The main advantage of TP is that parameter, gradient, and optimizer-state memory, along with per-device FLOPs, all decrease in proportion to the TP degree. However, TP requires very high communication volume (see Section~\ref{sec:tensor-parallel-theory}), which effectively restricts it to high-bandwidth intra-node communication fabrics such as AMD Infinity Fabric or NVIDIA NVLink. It is also difficult to overlap TP computation with communication, which makes efficient TP designs challenging. Finally, activations remain replicated across the TP group, which becomes the primary bottleneck on long sequences. Worse, the already-high TP communication volume is proportional to activation size and therefore grows with sequence length (see Section~\ref{sec:comm-mem-model}).

\subsection{Sequence Parallelism}
\label{sec:background-sp}

Sequence parallelism partitions the input along the sequence dimension while keeping model parameters replicated. Each rank processes only a subset of tokens, which reduces the per-rank activation memory and FLOPs. For SP on attention, each rank holds a shard of the sequence, computes local queries, keys, and values, and then exchanges the sequence-dependent tensors needed to recover the correct causal attention context.

For causal attention, naive equal-sized sequence chunks lead to load imbalance because later tokens attend to larger prefixes and therefore incur more work. A common remedy is a zigzag context partition in which each rank receives two disjoint subsequences, one near the beginning of the sequence and one near the end, so that the total causal attention workload is more evenly distributed. In this regime, SP reduces activation memory and attention FLOPs per rank, but leaves parameter memory fully replicated. It is therefore most attractive when activation storage, rather than model-state storage, is the dominant bottleneck. Sequence parallelism is becoming increasingly necessary as the supported context sizes for models increase dramatically. 

SP implementations differ in how the K/V exchange is realized. Ring Attention~\citep{liu2023ringattention} circulates K/V shards point-to-point around the SP group, overlapping each step's transfer with the corresponding blockwise FlashAttention computation. This keeps activation memory bounded to a single block at a time but requires $D{-}1$ sequential P2P steps per layer. An all-gather variant exchanges all K/V shards in a single collective and then runs FlashAttention once against the full-sequence keys and values, trading a higher peak K/V buffer for a single bandwidth-optimal collective and a simpler kernel call. DeepSpeed-Ulysses~\citep{jacobs2023deepspeedulysses} instead replaces the sequence-dimension exchange with all-to-alls that redistribute $Q/K/V$ across the head dimension, so each rank attends over the full sequence for a subset of heads. The forward-pass communication volume is comparable across these variants, but their sensitivity to topology and overlap differs. Ring is favorable when per-step compute is large enough to hide each P2P transfer and the fabric does not deliver full bandwidth to small subgroups. All-gather is favorable when the full collective fits in memory and the fabric is bandwidth-optimal at full subgroup size. All-to-all is favorable when the head count is large relative to the sequence shard size.

\subsection{Conventional TP+SP}
\label{sec:background-tpsp}
In prior works, TP and SP are combined by assigning them to orthogonal dimensions of the device mesh. If the TP degree is $T$ and the SP degree is $\Sigma$, then the model replica uses a total of $T\cdot\Sigma$ devices. TP reduces weight, optimizer, and gradient memory by a factor of $T$, while SP reduces activation memory by a factor of $\Sigma$. It also cleanly separates the communication patterns of the two strategies, with TP using reductions over the tensor-parallel group and SP using sequence-dependent communication operations over the sequence-parallel group.

However, this separation has two disadvantages: First, the communication domains of TP and SP may not map cleanly onto the cluster's communication topology. A configuration that is reasonable from a logical parallelism standpoint may still force one of the parallel groups to span slower inter-node links rather than high-bandwidth intra-node links. Second, a two-dimensional model replica mesh consumes many ranks that could be used to expand along the data-parallel axis.

\subsection{Parallelism Folding}
\label{sec:background-folding}

Parallelism folding is the idea of collapsing two logically distinct sharding dimensions onto a single device-mesh axis, so that each rank simultaneously owns shards along both partitioning schemes. Concretely, a fold of TP and SP gives every rank both a slice of the layer weights and a slice of the activations, rather than placing those two tensors on orthogonal axes of size $T$ and $\Sigma$.

While this lowers peak memory since more tensors are sharded across the same number of ranks instead of replicated, the trade-off is that each layer must now communicate along both folded axes during a single pass, where the two axes would otherwise communicate independently and at different points. This generally raises the per-layer communication volume, but this does not necessarily lead to a proportional increase in wall-clock cost if this communication is overlapped with compute (see Section \ref{sec:design}). 

Folding strategies also help to free ranks from the model-parallel mesh onto the data-parallel axis. This allows us to add DP replicas with a fixed cluster size. Such parallelism folding strategies are preferable only when its memory savings or reduction in redundant computation outweigh the additional communication cost, which is often achievable when sharding communication is across intra-node interconnects or overlapped with compute.

\section{Design}
\label{sec:design}

\begin{figure*}[hbtp]
    \centering
    \includegraphics[width=0.70\textwidth]{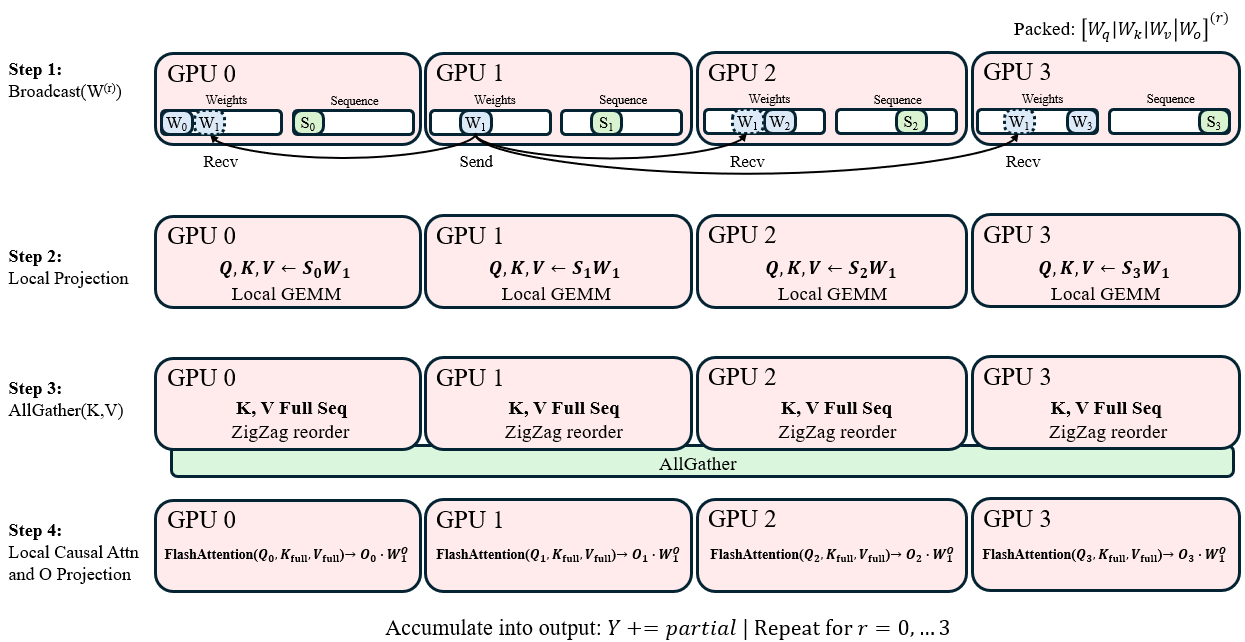}
    \caption{TSP Attention block design}
    \label{fig:tsp-attention}
\end{figure*}

\begin{figure*}[hbtp]
    \centering
    \includegraphics[width=0.70\textwidth]{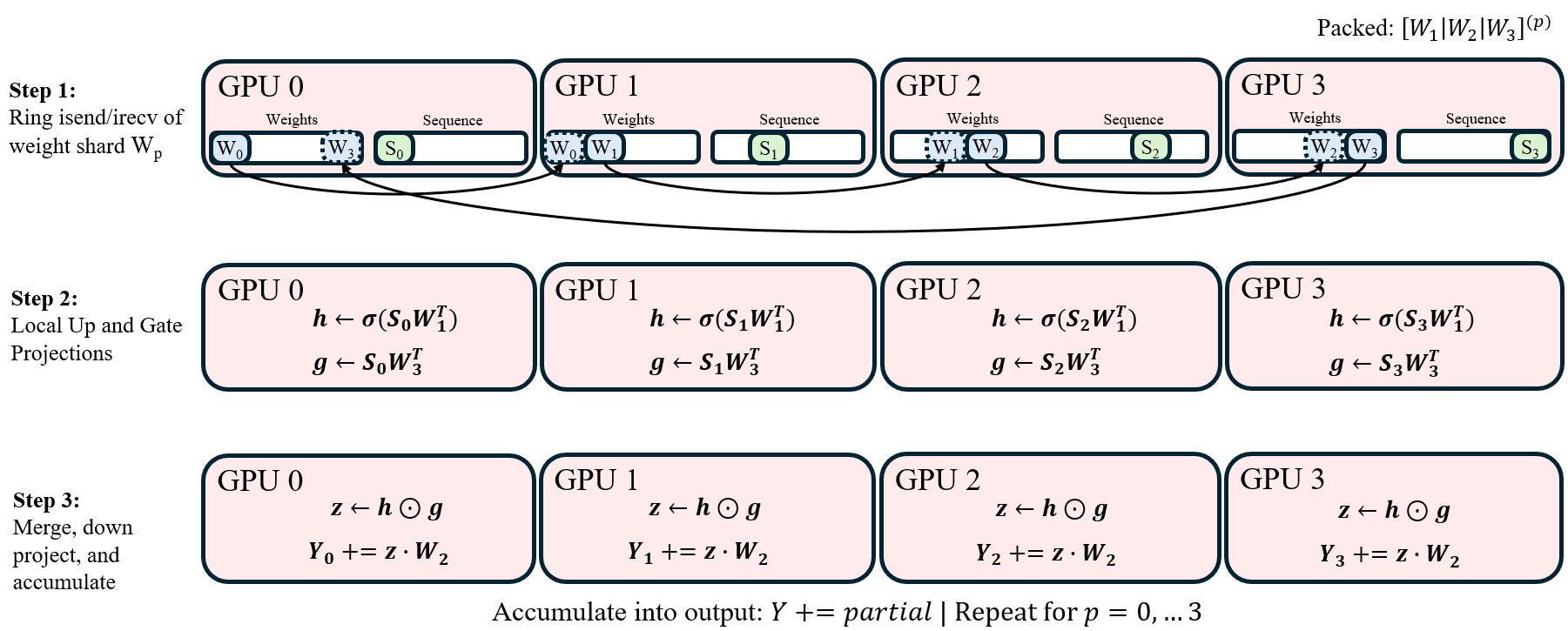}
    \caption{TSP MLP block design}
    \label{fig:tsp-mlp}
\end{figure*}

\subsection{TSP schedules}
\label{sec:algorithms}

We now describe the forward-pass designs for the TSP attention and gated MLP blocks. In both cases, each rank owns a shard of the sequence and a shard of the layer weights. TSP attention is illustrated in Figure~\ref{fig:tsp-attention} and specified in Algorithm~\ref{alg:tsp-mha}. Each rank stores a zigzag-partitioned sequence shard~\citep{brandon2023stripedattention} and a shard of the attention projections. The shards for $W_Q$, $W_K$, $W_V$, and $W_O$ are packed into a single buffer and broadcast one weight-owning rank at a time. After receiving the shard for iteration $r$, every rank applies it to its local tokens to produce the queries, keys, and values for that head shard. Causal attention still requires access to the full key/value context, so the local $K,V$ tensors are all-gathered across the TSP group and reordered back into sequence order before FlashAttention is applied. The resulting partial attention output is projected by the corresponding slice of $W_O$ and accumulated into the local output buffer. After all $D$ weight shards have been visited, each rank holds the full attention output for its local tokens.

The MLP schedule, illustrated in Figure~\ref{fig:tsp-mlp} and specified in Algorithm~\ref{alg:tsp-ring-mlp}, uses weight rotation instead of broadcast. Each rank begins with its local shards of the gate, up, and down projections, computes the gated MLP contribution for its local sequence tokens, and accumulates the result into a local output buffer. The weight shards then circulate around the TSP group using point-to-point communication, so that after $D$ iterations every rank has applied every MLP shard to its own sequence shard. Unlike standard tensor parallelism, this schedule does not require an all-reduce over the $BSh$-shaped activation output: the sequence remains local, and only the weights move. Thus the MLP communication cost is parameter-proportional, and the output accumulation remains local to each rank.

The use of broadcast for attention and a ring for the MLP is motivated by (but not limited to) AMD's intra-node topology. Unlike NVSwitch systems, where any subset of GPUs can communicate at full bandwidth, MI300X nodes use xGMI Infinity Fabric, whose bandwidth is highest when all eight GPUs participate in a collective. A $k$-GPU subgroup has aggregate bandwidth that scales as $(k-1) \cdot B_{\mathrm{link}}$ rather than a topology-wide constant $B_{\max}$~\citep{anthony2025trainingfoundationmodelsfullstack}, which makes bandwidth-bound collectives over fewer ranks less attractive. The opposite trend holds for small-message latency, where fewer participants is better. In TSP, the MLP point-to-point ring is efficient because each step is compute-heavy enough to hide the transfer: the moved shard has size $P_{\mathrm{mlp}}\beta_P/D$ while the matching GEMMs cost on the order of $BS \cdot (Fh/D) \cdot h$. Attention instead uses a broadcast loop because its dominant communication is the sequence-wise K/V all-gather, an activation-proportional collective that benefits from engaging all ranks simultaneously. The packed weight broadcast is smaller than the K/V exchange and is amortized over the full TSP group. Both primitives are thus matched to the arithmetic intensity and bandwidth structure of the corresponding block, which matters especially on xGMI-based AMD systems and is no worse on fully symmetric NVSwitch fabrics.

\begin{algorithm*}
\caption{TSP Multi-Head Attention (forward pass on rank $p$). For clarity we present the MHA case; GQA replaces the K/V shard widths $h/D$ with $h/(gD)$ and reduces the K/V all-gather volume by $1/g$ accordingly.}
\label{alg:tsp-mha}
\begin{algorithmic}[1]
\Require $\mathbf{X} \in \mathbb{R}^{B \times S_{\mathrm{loc}} \times h}$ \Comment{zigzag-partitioned local sequence shard, $S_{\mathrm{loc}} = S/D$}
\Require Local weight shards $\mathbf{W}_Q^{(p)}, \mathbf{W}_K^{(p)}, \mathbf{W}_V^{(p)} \in \mathbb{R}^{(h/D) \times h}$, \; $\mathbf{W}_O^{(p)} \in \mathbb{R}^{h \times (h/D)}$
\Require Head bucket size $B_h$, world size $D$, head dim $d$, heads per shard $H_s = n_h/D$
\Ensure $\mathbf{Y} \in \mathbb{R}^{B \times S_{\mathrm{loc}} \times h}$

\State $S_c \gets S_{\mathrm{loc}} / 2$ \Comment{half local seq length (zigzag)}
\State $\mathbf{Y} \gets \mathbf{0}_{B \times S_{\mathrm{loc}} \times h}$

\For{$r = 0, \ldots, D-1$} \Comment{loop over weight-owning ranks}
    \State $\mathbf{W} \gets \Call{Broadcast}{[\mathbf{W}_Q^{(p)} \| \mathbf{W}_K^{(p)} \| \mathbf{W}_V^{(p)} \| {\mathbf{W}_O^{(p)}}^\top],\; \text{src}=r}$ \Comment{packed bcast}
    \State Unpack $\mathbf{W}_Q^{(r)},\, \mathbf{W}_K^{(r)},\, \mathbf{W}_V^{(r)},\, {\mathbf{W}_O^{(r)}}^\top$ from $\mathbf{W}$
    \State $\mathbf{Q} \gets \mathbf{X}\,{\mathbf{W}_Q^{(r)}}^\top$, \;
           $\mathbf{K} \gets \mathbf{X}\,{\mathbf{W}_K^{(r)}}^\top$, \;
           $\mathbf{V} \gets \mathbf{X}\,{\mathbf{W}_V^{(r)}}^\top$
           \Comment{local projections}
    \State Reshape $\mathbf{Q}, \mathbf{K}, \mathbf{V}$ to $[B,\, S_{\mathrm{loc}},\, H_s,\, d]$
    \State $s_1 \gets (p+1)\,S_c$, \; $s_2 \gets (2D - p)\,S_c$ \Comment{zigzag causal bounds}
    \State $\mathbf{A} \gets \mathbf{0}_{B \times S_{\mathrm{loc}} \times H_s \times d}$

    \For{$b = 0,\, B_h,\, 2B_h,\, \ldots,\, H_s - B_h$} \Comment{head bucketing}
        \State $\mathbf{K}_b,\, \mathbf{V}_b \gets \mathbf{K}_{[:,:,\,b:b+B_h,\,:]}$, \; $\mathbf{V}_{[:,:,\,b:b+B_h,\,:]}$
        \State $[\hat{\mathbf{K}}_b \| \hat{\mathbf{V}}_b] \gets \Call{AllGather}{[\mathbf{K}_b \| \mathbf{V}_b],\; \text{dim}=\text{seq}}$
        \State $\hat{\mathbf{K}}_b \gets \Call{ZigZagReorder}{\hat{\mathbf{K}}_b}$, \;
               $\hat{\mathbf{V}}_b \gets \Call{ZigZagReorder}{\hat{\mathbf{V}}_b}$
        \State $\mathbf{Q}_b^{(1)},\, \mathbf{Q}_b^{(2)} \gets \Call{Split}{\mathbf{Q}_{[:,:,\,b:b+B_h,\,:]},\; \text{dim}=\text{seq}}$
        \State $\mathbf{O}_b^{(1)} \gets \Call{CausalAttn}{\mathbf{Q}_b^{(1)},\, \hat{\mathbf{K}}_{b}^{[:,:s_1]},\, \hat{\mathbf{V}}_{b}^{[:,:s_1]}}$
        \State $\mathbf{O}_b^{(2)} \gets \Call{CausalAttn}{\mathbf{Q}_b^{(2)},\, \hat{\mathbf{K}}_{b}^{[:,:s_2]},\, \hat{\mathbf{V}}_{b}^{[:,:s_2]}}$
        \State $\mathbf{A}_{[:,:,\,b:b+B_h,\,:]} \gets [\mathbf{O}_b^{(1)};\; \mathbf{O}_b^{(2)}]$ \Comment{concat along seq}
    \EndFor

    \State $\mathbf{Y} \gets \mathbf{Y} + \Call{Flatten}{\mathbf{A}}\, {\mathbf{W}_O^{(r)}}^\top$ \Comment{output projection + accumulate}
\EndFor

\State \Return $\mathbf{Y}$
\end{algorithmic}
\end{algorithm*}

\begin{algorithm*}
\caption{TSP Gated Ring MLP (forward pass on rank $p$)}
\label{alg:tsp-ring-mlp}
\begin{algorithmic}[1]
\Require $\mathbf{X} \in \mathbb{R}^{B \times S_{\mathrm{loc}} \times h}$ \Comment{local sequence shard on rank $p$, $S_{\mathrm{loc}} = S/D$}
\Require Local weight shards $\mathbf{W}_1^{(p)},\, \mathbf{W}_3^{(p)} \in \mathbb{R}^{(Fh/D) \times h}$, \; $\mathbf{W}_2^{(p)} \in \mathbb{R}^{(Fh/D) \times h}$
\Require Activation $\sigma$, world size $D$
\Ensure $\mathbf{Y} \in \mathbb{R}^{B \times S_{\mathrm{loc}} \times h}$

\State $\mathbf{Y} \gets \mathbf{0}_{B \times S_{\mathrm{loc}} \times h}$
\State $\mathbf{W}_1,\, \mathbf{W}_2,\, \mathbf{W}_3 \gets \mathbf{W}_1^{(p)},\, \mathbf{W}_2^{(p)},\, \mathbf{W}_3^{(p)}$

\For{$t = 0$ \textbf{to} $D - 1$}
    \If{$t < D - 1$} \Comment{overlap P2P with compute below}
        \State $\mathbf{W}_1',\, \mathbf{W}_2',\, \mathbf{W}_3' \gets \Call{AsyncRingSendRecv}{\mathbf{W}_1,\, \mathbf{W}_2,\, \mathbf{W}_3}$
        \Statex \hspace{\algorithmicindent}\hspace{\algorithmicindent}\Comment{send to $(p{+}1) \bmod D$, recv from $(p{-}1) \bmod D$}
    \EndIf

    \State $\mathbf{G} \gets \sigma\!\bigl(\mathbf{X}\,\mathbf{W}_1^\top\bigr)$ \Comment{gate projection + activation}
    \State $\mathbf{H} \gets \mathbf{X}\,\mathbf{W}_3^\top$ \Comment{up-projection}
    \State $\mathbf{Y} \gets \mathbf{Y} + (\mathbf{G} \odot \mathbf{H})\,\mathbf{W}_2$ \Comment{down-projection + accumulate}

    \If{$t < D - 1$}
        \State \Call{Wait}{\,}
        \State $\mathbf{W}_1,\, \mathbf{W}_2,\, \mathbf{W}_3 \gets \mathbf{W}_1',\, \mathbf{W}_2',\, \mathbf{W}_3'$
    \EndIf
\EndFor

\State \Return $\mathbf{Y}$
\end{algorithmic}
\end{algorithm*}

\begin{figure*}[hbtp]
    \centering
    \includegraphics[width=\textwidth]{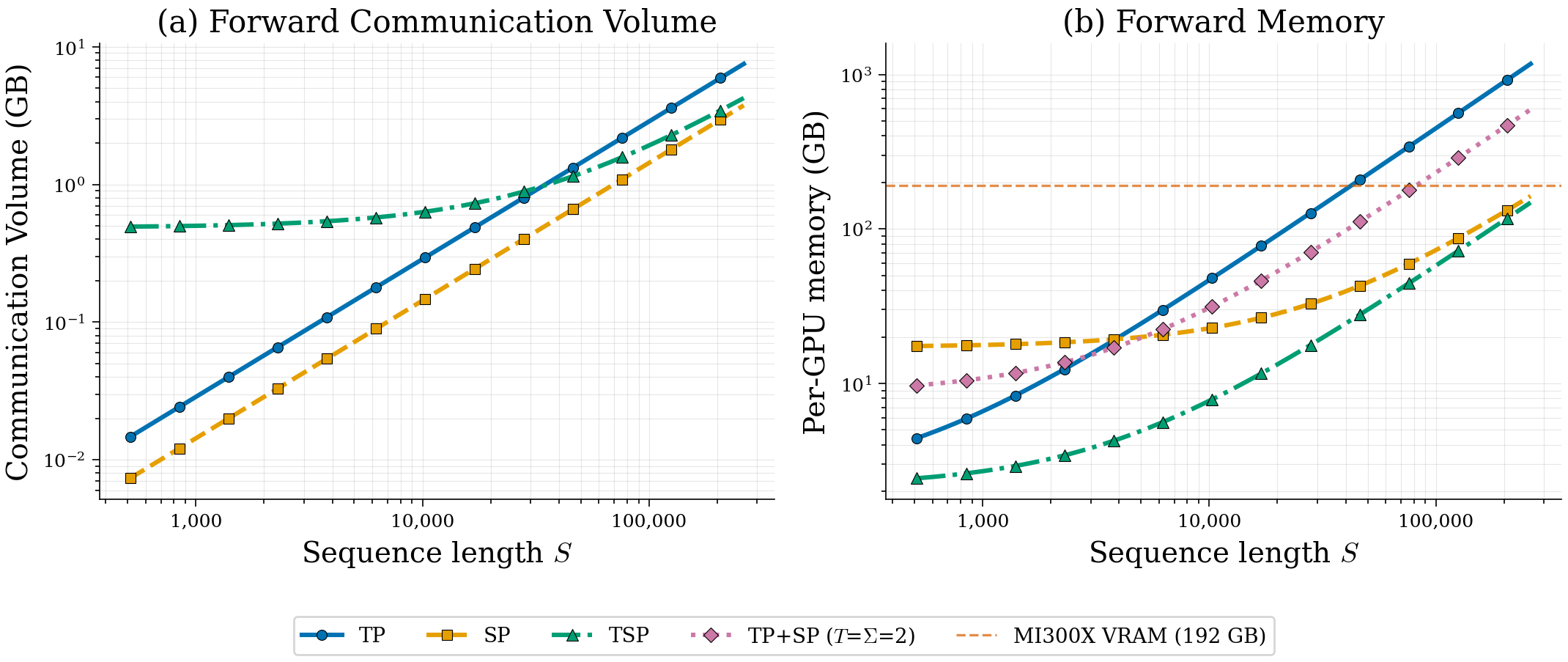}
    \caption{Forward-pass theoretical communication volume~(a) and per-GPU memory~(b) as a function of sequence length, for the 7B reference model used throughout the paper ($h{=}4096$, $L{=}32$, MHA, $F{=}4$; full configuration in Appendix~\ref{app:model-config}). Parallelism degree $D{=}8$, micro-batch $B{=}1$. TSP communication exceeds TP at short sequences due to weight movement, but the gap narrows as the activation-proportional term grows.}
    \label{fig:comm-mem-vs-seqlen}
\end{figure*}

\begin{figure*}[hbtp]
    \centering
    \includegraphics[width=\textwidth]{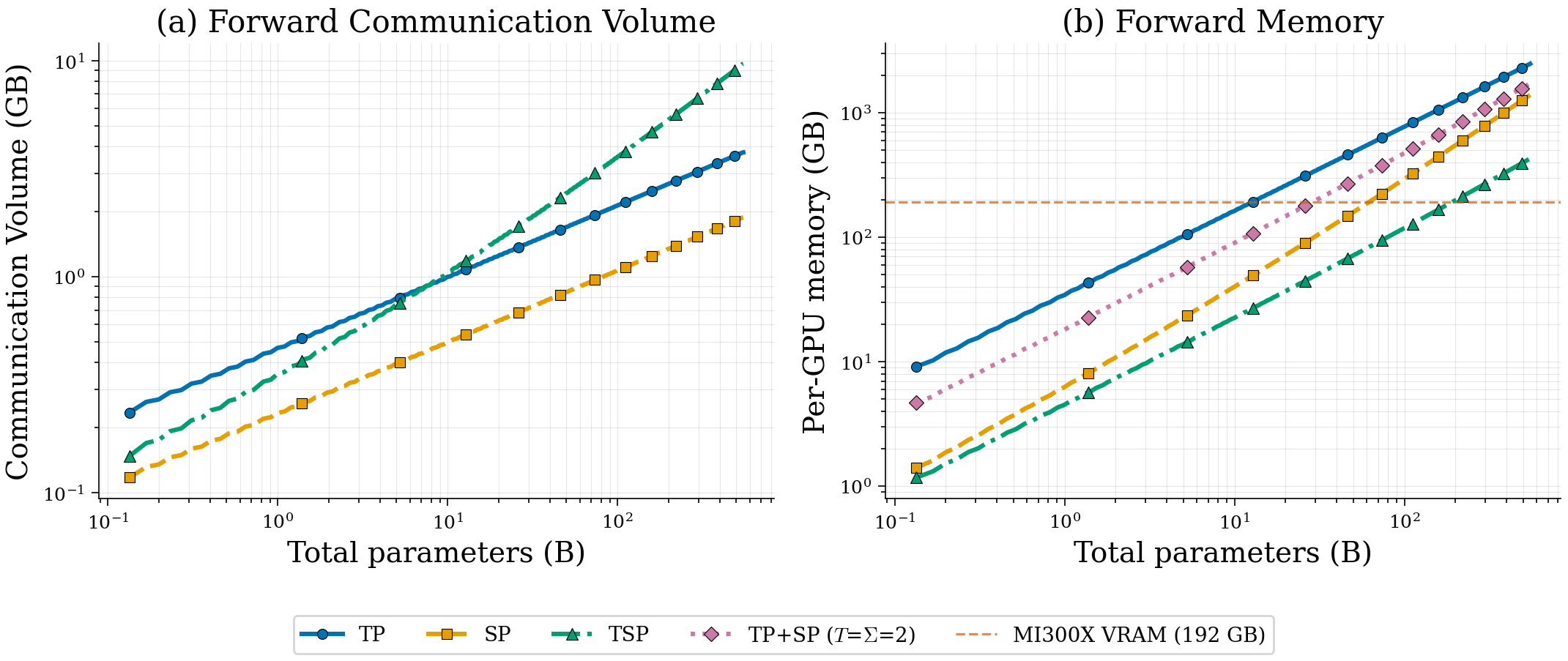}
    \caption{Forward-pass theoretical communication volume~(a) and per-GPU memory~(b) as a
    function of total parameter count ($S{=}32768$, $D{=}8$, $B{=}1$, MHA).
    TSP provides the lowest per-GPU memory across the full range, which translates to throughput improvements due to reduced parallelism and higher batch sizes.}
    \label{fig:comm-mem-vs-params}
\end{figure*}

\begin{figure}[hbtp]
    \centering
    \includegraphics[width=1.04\columnwidth]{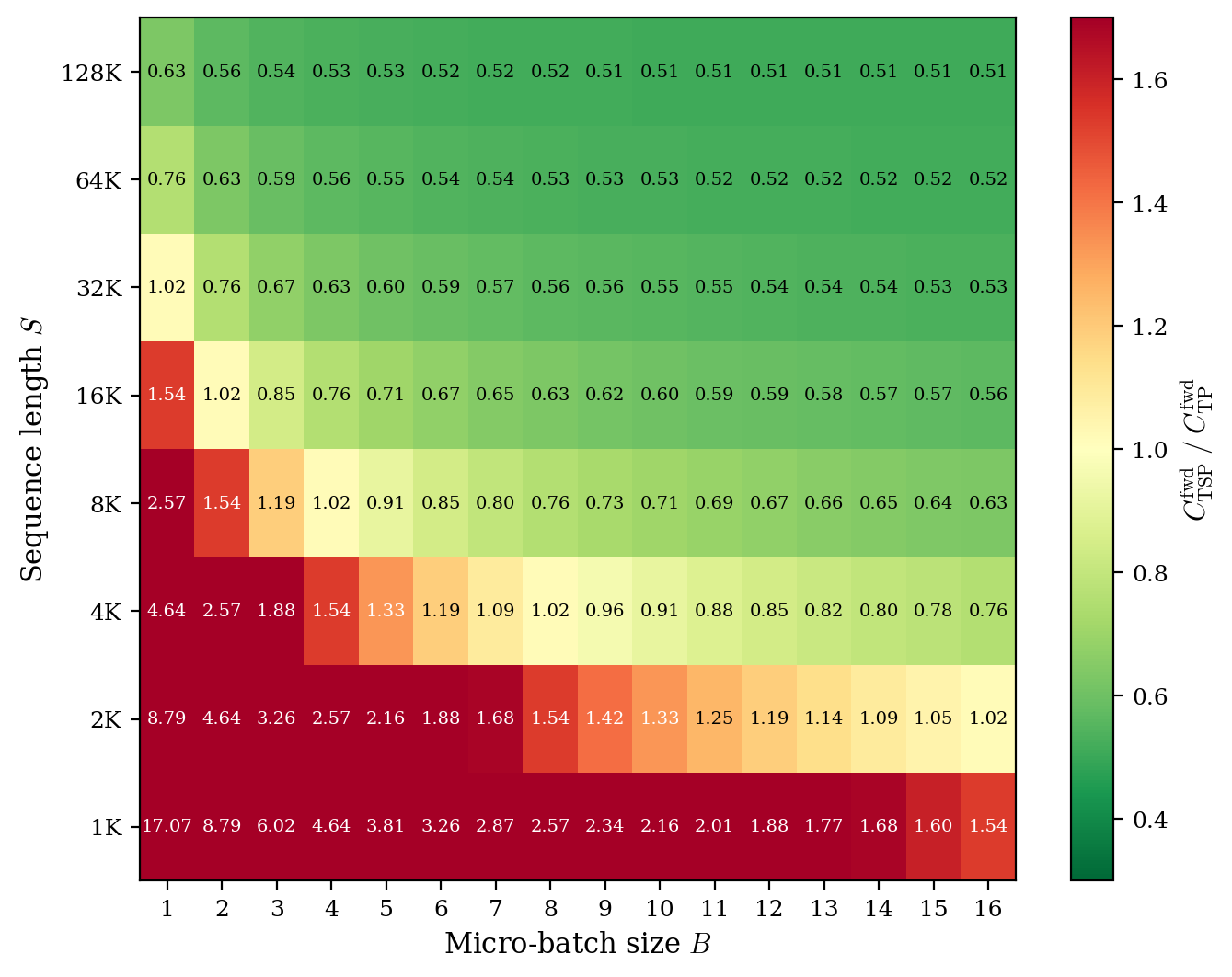}
    \caption{Ratio of theoretical TSP to TP forward communication volume as a function of
    micro-batch size $B$ and sequence length $S$ ($h{=}4096$, $D{=}8$, MHA).
    Green cells (ratio ${<}\,1$) indicate regimes where TSP communicates less
    than TP. The break-even boundary corresponds to the condition $BS > 8h$.}
    \label{fig:tsp-tp-crossover}
\end{figure}

\section{Theoretical Analysis}
\label{sec:comm-mem-model}

In this section, we build a detailed theoretical model of the memory costs, FLOPs, and communication volume of different parallelism schemes. We compare data parallelism (DP), tensor parallelism (TP), sequence parallelism (SP), conventional two-dimensional tensor--sequence parallelism (TP+SP), and tensor--sequence parallelism on a single folded axis (TSP). Throughout, we use the notation in Table~\ref{tab:tsp_notation}.

We assume a dense decoder-only transformer with pre-norm residual connections, multi-head (or grouped-query) attention with learned linear projections for $Q$, $K$, $V$, and $O$, and a gated MLP (e.g.\ SwiGLU) with three weight matrices. We ignore the parameter and FLOP contributions of the embedding/unembedding layers, positional encodings, and layer normalization, and assume models that are large enough to make these architectural choices unimportant.

\begin{table}[t]
\centering
\caption{Notation used in \S\ref{sec:comm-mem-model}.}
\label{tab:tsp_notation}
\begin{tabularx}{\linewidth}{@{}cX@{}}
\toprule
\textbf{Symbol} & \textbf{Definition} \\
\midrule
$D$ & Parallel degree; number of devices (GPUs) in the parallel group \\
$T,\Sigma$ & Tensor- and sequence-parallel degrees in a 2-dimensional mesh, with $D=T\Sigma$ \\
$B$ & Micro-batch size per GPU \\
$S$ & Sequence length \\
$h$ & Model embedding / residual dimension \\
$n_h,\;n_{kv}$ & Number of query heads and key/value heads \\
$d$ & Per-head dimension, $d=h/n_h$ \\
$g$ & GQA ratio, $g=n_h/n_{kv}$ \\
$F$ & FFN expansion factor \\
$L$ & Number of transformer layers \\
$\beta_P$ & Bytes per parameter element (forward precision) \\
$\beta_g$ & Bytes per gradient element \\
$n_o$ & Number of optimizer states (e.g.\ 3 for mixed-precision AdamW: first moment, second moment, and master parameter copy) \\
$\beta_o$ & Bytes per optimizer-state element (e.g.\ 4 for fp32 states) \\
$P_{\mathrm{attn}}$ & Attention parameters per layer, $2\!\left(1+\frac{1}{g}\right)h^2$ \\
$P_{\mathrm{mlp}}$ & Gated-MLP parameters per layer, $3Fh^2$ \\
$P_L$ & Total parameters per layer, $P_{\mathrm{attn}}+P_{\mathrm{mlp}}$ \\
$C$ & Communication volume (bytes) \\
\bottomrule
\end{tabularx}
\end{table}

For one transformer layer, the parameter count is
\begin{equation}
P_L
=
P_{\mathrm{attn}} + P_{\mathrm{mlp}}
=
\left(2\!\left(1+\frac{1}{g}\right)+3F\right)h^2.
\end{equation}
The attention term accounts for the $Q$, $K$, $V$, and $O$ projection matrices (with $K$ and $V$ projecting to $h/g$ dimensions each under grouped-query attention). The MLP term accounts for the gate, up, and down projections in a gated MLP.

\subsection{Memory decomposition}
We decompose the per-GPU persistent memory into four terms:
\begin{enumerate}
\item \textbf{Parameter memory ($M_p$):} the storage for the model weights, $LP_L\beta_P$ for the full model.
\item \textbf{Gradient memory ($M_g$):} the storage for gradient accumulators, $LP_L\beta_g$ for the full model.
\item \textbf{Optimizer state memory ($M_o$):} the storage for $n_o$ optimizer states at $\beta_o$ bytes each (e.g.\ first and second moments plus a high-precision parameter copy for Adam), totalling $LP_Ln_o\beta_o$ for the full model.
\item \textbf{Activation memory ($M_a$):} the storage for intermediate activations needed for backpropagation.
\end{enumerate}
Parameter, gradient, and optimizer memory are proportional to the parameter count, so their sum is $LP_L(\beta_P + \beta_g + n_o\beta_o)$.

\subsection{Activation memory}
Following~\cite{korthikanti2023reducing}, we distinguish two activation-recomputation regimes. Under selective recomputation, most activations are stored but the attention probability matrices and associated dropout masks are recomputed during the backward pass, giving
\begin{equation}
\label{eq:act_sel}
M_a^{\mathrm{sel}}
\;=\;
LBSh\!\left(16\beta_P + 2\right).
\end{equation}
Under full recomputation, only the input to each transformer layer is stored and all other activations are recomputed from these checkpoints, giving
\begin{equation}
\label{eq:act_full}
M_a^{\mathrm{full}}
\;=\;
LBSh\,\beta_P.
\end{equation}
When no recomputation is used, all activations, including the attention probability matrices, are retained:
\begin{equation}
\label{eq:act_none}
M_a^{\mathrm{none}}
\;=\;
LBSh\!\left(16\beta_P + 2 + \left(2\beta_P+1\right)\frac{n_h S}{h}\right).
\end{equation}

\subsection{Communication volume}
We denote by $C$ the communication volume in bytes transferred per layer per device. For a message of total size $N$, ring-style collectives incur the following communication volume:
\begin{align*}
\text{all-reduce:}\quad & 2N\,\tfrac{D-1}{D}, \\
\text{all-gather:}\quad & N\,\tfrac{D-1}{D}, \\
\text{broadcast:}\quad & N.
\end{align*}
Importantly, activation recomputation introduces additional communication overhead: when activations are recomputed during the backward pass, the forward-pass communication for that layer must be repeated. We account for this in the training totals below by writing $C^{\mathrm{fwd+bwd}}$ for training without recomputation and $C^{\mathrm{fwd+bwd+recomp}}$ for training with full recomputation, where the latter incurs an extra forward pass of communication per layer.

\subsubsection{Data Parallelism (DP)}

DP replicates the full model on every device and shards only the batch. The forward pass requires no communication; the backward pass synchronizes gradients via a single all-reduce of size $P_L\beta_g$:
\begin{align}
C_{\mathrm{DP}}^{\mathrm{fwd}} &= 0, \\
C_{\mathrm{DP}}^{\mathrm{fwd+bwd}} &= 2P_L\beta_g\,\tfrac{D-1}{D}, \\
C_{\mathrm{DP}}^{\mathrm{fwd+bwd+recomp}} &= 2P_L\beta_g\,\tfrac{D-1}{D}.
\end{align}
Because the forward pass is communication-free, recomputation adds no communication overhead. Every device stores the full model, so no weight-proportional memory term is reduced:
\begin{align*}
M_P &= LP_L\beta_P, & M_g &= LP_L\beta_g, \\
M_o &= LP_Ln_o\beta_o, & M_a &\in \{M_a^{\mathrm{full}}, M_a^{\mathrm{sel}}, M_a^{\mathrm{none}}\}.
\end{align*}

\subsubsection{Tensor Parallelism (TP)}
\label{sec:tensor-parallel-theory}

TP shards weights across $D$ devices while each device still processes the full sequence. In the implementation considered here, attention is head-parallel and the MLP is column-parallel followed by row-parallel reduction. The attention output projection requires one all-reduce of size $BSh\beta_P$, and the MLP output requires another; the backward pass introduces two more all-reduces of the same size. Recomputation (but only full block-wise recomputation, since selective recomputation in the style of \cite{korthikanti2023reducing} can target only the local activations per rank within the attention and MLP blocks) repeats the two forward all-reduces:
\begin{align}
C_{\mathrm{TP}}^{\mathrm{fwd}} &= 4BSh\beta_P\,\tfrac{D-1}{D}, \\
C_{\mathrm{TP}}^{\mathrm{fwd+bwd}} &= 8BSh\beta_P\,\tfrac{D-1}{D}, \\
C_{\mathrm{TP}}^{\mathrm{fwd+bwd+full\_recomp}} &= 12BSh\beta_P\,\tfrac{D-1}{D}.
\end{align}
TP communication is therefore activation-proportional rather than weight-proportional.

Because parameters, gradients, and optimizer states are sharded across $D$ devices while each device retains the full sequence:
\begin{align*}
M_P &= \tfrac{LP_L\beta_P}{D}, & M_g &= \tfrac{LP_L\beta_g}{D}, \\
M_o &= \tfrac{LP_Ln_o\beta_o}{D}, & M_a &\in \{M_a^{\mathrm{full}}, M_a^{\mathrm{sel}}, M_a^{\mathrm{none}}\}.
\end{align*}
TP reduces all weight-proportional memory in the attention and MLP blocks by $1/D$, but leaves activations unsharded.

\subsubsection{Sequence Parallelism (SP)}

SP replicates weights and shards the activations across $D$ devices. In the implementation considered here, the sequence is partitioned using a ZigZag layout for causal load balancing. The MLP remains local because weights are replicated, and only attention communicates.

Of the SP variants surveyed in \S\ref{sec:background-sp}, we analyze the all-gather formulation, since it uses the same K/V exchange primitive as the TSP attention schedule (\S\ref{sec:algorithms}) and therefore yields a like-for-like cost comparison; ring and all-to-all variants have comparable forward volume but differ in overlap and topology sensitivity. Each device computes local $Q,K,V$ on its sequence shard and all-gathers $K,V$ across the SP group. The forward all-gather moves two tensors of total size $BSh\beta_P/g$; the backward pass repeats this exchange for the attention gradients and adds a gradient all-reduce for the replicated weights. Recomputation repeats the forward K/V exchange:
\begin{align}
C_{\mathrm{SP}}^{\mathrm{fwd}} &= \frac{2BSh\beta_P}{g}\,\tfrac{D-1}{D}, \\
C_{\mathrm{SP}}^{\mathrm{fwd+bwd}} &= \frac{4BSh\beta_P}{g}\,\tfrac{D-1}{D} + 2P_L\beta_g\,\tfrac{D-1}{D}, \\
C_{\mathrm{SP}}^{\mathrm{fwd+bwd+recomp}} &= \frac{6BSh\beta_P}{g}\,\tfrac{D-1}{D} + 2P_L\beta_g\,\tfrac{D-1}{D}.
\end{align}

For MHA ($g=1$), the forward cost reduces to $2BSh\beta_P\,\tfrac{D-1}{D}$.

Because the full model is replicated, all weight-proportional memory is unchanged; only activations are sharded by $D$:
\begin{align*}
M_P &= LP_L\beta_P, & M_g &= LP_L\beta_g, \\
M_o &= LP_Ln_o\beta_o, & M_a &\in \{M_a^{\mathrm{full}}/D, M_a^{\mathrm{sel}}/D, M_a^{\mathrm{none}}/D\}.
\end{align*}

\subsubsection{Tensor and Sequence Parallelism (TSP)}

TSP shards both weights and sequence on the same $D$-way axis. Each device stores only $1/D$ of the model parameters and only $1/D$ of the sequence, but must move weights at runtime.

For attention, the implementation loops over the $D$ weight shards. In each iteration, it broadcasts a packed shard containing $W_q$, $W_k$, $W_v$, and $W_o^\top$ (costing $P_{\mathrm{attn}}\beta_P/D$ per broadcast, totalling $P_{\mathrm{attn}}\beta_P$ over all $D$ iterations), then all-gathers the corresponding K/V activations across the sharded sequence dimension. For the gated MLP, the ring algorithm rotates $W_1$, $W_2$, and $W_3$ around the parallel group, where each of the $D-1$ ring steps transfers a shard of size $P_{\mathrm{mlp}}\beta_P/D$. Combining these terms, the forward communication per layer is
\begin{equation}
C_{\mathrm{TSP}}^{\mathrm{fwd}}
= P_{\mathrm{attn}}\beta_P + P_{\mathrm{mlp}}\beta_P\,\tfrac{D-1}{D} + \frac{2BSh\beta_P}{g}\,\tfrac{D-1}{D}.
\end{equation}
The backward pass mirrors the forward weight movement and K/V exchange. Like DP and SP, TSP requires a separate full-model gradient all-reduce. Recomputation repeats the forward communication:
\begin{align}
C_{\mathrm{TSP}}^{\mathrm{fwd+bwd}}
&= 2P_{\mathrm{attn}}\beta_P + 2P_{\mathrm{mlp}}\beta_P\,\tfrac{D-1}{D} \notag \\
&\quad + \frac{4BSh\beta_P}{g}\,\tfrac{D-1}{D} + P_L\beta_g\,\tfrac{D-1}{D},
\end{align}
\begin{align}
C_{\mathrm{TSP}}^{\mathrm{fwd+bwd+recomp}}
&= 3P_{\mathrm{attn}}\beta_P + 3P_{\mathrm{mlp}}\beta_P\,\tfrac{D-1}{D} \notag \\
&\quad + \frac{6BSh\beta_P}{g}\,\tfrac{D-1}{D} + 2P_L\beta_g\,\tfrac{D-1}{D}.
\end{align}

Because both weights and activations are sharded by $D$:
\begin{align*}
M_P &= \tfrac{LP_L\beta_P}{D}, & M_g &= \tfrac{LP_L\beta_g}{D}, \\
M_o &= \tfrac{LP_Ln_o\beta_o}{D}, & M_a &\in \{M_a^{\mathrm{full}}/D, M_a^{\mathrm{sel}}/D, M_a^{\mathrm{none}}/D\}.
\end{align*}
TSP is therefore the only scheme that reduces both weight-proportional and activation memory by the same $1/D$ factor on a single parallelism axis.

\subsubsection{Conventional TP+SP on a 2D Mesh}

For completeness, consider a two-dimensional mesh with TP degree $T$ and SP degree $\Sigma$, where $D=T\Sigma$. In this setting, weights are sharded by $T$ and sequence is sharded by $\Sigma$. The dominant forward communication combines SP K/V exchange (all-gathered over the $\Sigma$-way SP group) with TP output reductions (all-reduced over the $T$-way TP group):
\begin{equation}
C_{\mathrm{TP+SP}}^{\mathrm{fwd}}
= \frac{2BSh\beta_P}{gT}\,\tfrac{\Sigma-1}{\Sigma} + \frac{4BSh\beta_P}{\Sigma}\,\tfrac{T-1}{T}.
\end{equation}
Training doubles the activation terms. Because weights are sharded along the TP axis but replicated along the SP axis, parameter gradients must be all-reduced over the $\Sigma$-way SP group at cost $2(P_L\beta_g/T)\,\tfrac{\Sigma-1}{\Sigma}$:
\begin{align}
C_{\mathrm{TP+SP}}^{\mathrm{fwd+bwd}}
&= \frac{4BSh\beta_P}{gT}\,\tfrac{\Sigma-1}{\Sigma} + \frac{8BSh\beta_P}{\Sigma}\,\tfrac{T-1}{T} \notag \\
&\quad + \frac{2P_L\beta_g}{T}\,\tfrac{\Sigma-1}{\Sigma},
\end{align}
\begin{align}
C_{\mathrm{TP+SP}}^{\mathrm{fwd+bwd+recomp}}
&= \frac{6BSh\beta_P}{gT}\,\tfrac{\Sigma-1}{\Sigma} + \frac{12BSh\beta_P}{\Sigma}\,\tfrac{T-1}{T} \notag \\
&\quad + \frac{2P_L\beta_g}{T}\,\tfrac{\Sigma-1}{\Sigma}.
\end{align}
Weight-proportional memory scales with $1/T$, while activation memory scales with $1/\Sigma$:
\begin{align*}
M_P &= \tfrac{LP_L\beta_P}{T}, & M_g &= \tfrac{LP_L\beta_g}{T}, \\
M_o &= \tfrac{LP_Ln_o\beta_o}{T}, & M_a &\in \{M_a^{\mathrm{full}}/\Sigma, M_a^{\mathrm{sel}}/\Sigma, M_a^{\mathrm{none}}/\Sigma\}.
\end{align*}
Unlike TSP, TP+SP does not apply a single $D$-way sharding factor to both memory classes simultaneously.

\begin{table*}[t]
\centering
\caption{Dominant per-layer communication volume $C$ (bytes per device) in the forward pass.}
\label{tab:comm_summary}
\begin{tabular}{lccc}
\toprule
Strategy & Activation term & Weight term & Total forward volume \\
\midrule
DP & $0$ & $0$ & $0$ \\[2pt]
TP & $4BSh\beta_P\,\tfrac{D-1}{D}$ & $0$ & $4BSh\beta_P\,\tfrac{D-1}{D}$ \\[4pt]
SP & $\dfrac{2BSh\beta_P}{g}\,\tfrac{D-1}{D}$ & $0$ & $\dfrac{2BSh\beta_P}{g}\,\tfrac{D-1}{D}$ \\[6pt]
TP+SP & $\dfrac{2BSh\beta_P}{gT}\,\tfrac{\Sigma\!-\!1}{\Sigma} + \dfrac{4BSh\beta_P}{\Sigma}\,\tfrac{T\!-\!1}{T}$ & $0$ & same \\[6pt]
TSP & $\dfrac{2BSh\beta_P}{g}\,\tfrac{D-1}{D}$ & $P_{\mathrm{attn}}\beta_P + P_{\mathrm{mlp}}\beta_P\,\tfrac{D-1}{D}$ & sum \\
\bottomrule
\end{tabular}
\end{table*}

\begin{table*}[t]
\centering
\caption{Dominant per-GPU memory by category. $M_a^{\mathrm{full}}$, $M_a^{\mathrm{sel}}$, and $M_a^{\mathrm{none}}$ denote activation memory under full recomputation~\eqref{eq:act_full}, selective recomputation~\eqref{eq:act_sel}, and no recomputation~\eqref{eq:act_none}, respectively. Temporary communication buffers are omitted.}
\label{tab:mem_summary}
\begin{tabular}{lcccc}
\toprule
Strategy & Parameters & Gradients & Optimizer states & Activations \\
\midrule
DP     & $LP_L\beta_P$           & $LP_L\beta_g$           & $LP_Ln_o\beta_o$           & $M_a^{\{\cdot\}}$ \\[2pt]
TP     & $LP_L\beta_P/D$         & $LP_L\beta_g/D$         & $LP_Ln_o\beta_o/D$         & $M_a^{\{\cdot\}}$ \\[2pt]
SP     & $LP_L\beta_P$           & $LP_L\beta_g$           & $LP_Ln_o\beta_o$           & $M_a^{\{\cdot\}}/D$ \\[2pt]
TP+SP  & $LP_L\beta_P/T$         & $LP_L\beta_g/T$         & $LP_Ln_o\beta_o/T$         & $M_a^{\{\cdot\}}/\Sigma$ \\[2pt]
TSP    & $LP_L\beta_P/D$         & $LP_L\beta_g/D$         & $LP_Ln_o\beta_o/D$         & $M_a^{\{\cdot\}}/D$ \\
\bottomrule
\end{tabular}
\end{table*}

\subsection{FLOPs}
Ignoring softmax, normalization, masking, and other pointwise operations, the dominant forward FLOPs of a single transformer layer are
\begin{align}
F_{\mathrm{attn}}
&\approx
4\!\left(1+\frac{1}{g}\right)BSh^2 + 4BS^2h, \\
F_{\mathrm{mlp}}
&\approx
6FBSh^2,
\end{align}
where the first term in $F_{\mathrm{attn}}$ accounts for the $Q,K,V,O$ projections (with $K,V$ projecting to $h/g$ dimensions each) and the second for the attention score and context computation over all $n_h$ query heads. Thus
\begin{equation}
F_{\mathrm{layer}}
\;\approx\;
\left(4\!\left(1+\frac{1}{g}\right)+6F\right)BSh^2 + 4BS^2h.
\end{equation}
DP performs the full layer on every device:
\begin{equation}
F_{\mathrm{DP}} \approx F_{\mathrm{layer}}.
\end{equation}
By contrast, TP, SP, TP+SP, and TSP all ideally reduce the per-device leading-order arithmetic by the parallel degree:
\begin{align}
F_{\mathrm{TP}}
\;\approx\;
F_{\mathrm{SP}}
\;\approx\;
F_{\mathrm{TSP}}
&\;\approx\;
\frac{F_{\mathrm{layer}}}{D}, \\
F_{\mathrm{TP+SP}}
&\;\approx\;
\frac{F_{\mathrm{layer}}}{T\Sigma}.
\end{align}

\begin{table}[t]
\centering
\caption{Leading-order forward FLOPs per device, ignoring softmax, normalization, and other pointwise operations. Backward FLOPs are of the same order.}
\label{tab:flop_summary}
\setlength{\tabcolsep}{3pt}
\small
\begin{tabularx}{\linewidth}{@{}lX@{}}
\toprule
Strategy & Forward FLOPs per device \\
\midrule
DP & $\left(4\!\left(1+\tfrac{1}{g}\right)+6F\right)BSh^2 + 4BS^2h$ \\[4pt]
TP, SP, TSP & $\left[\left(4\!\left(1+\tfrac{1}{g}\right)+6F\right)BSh^2 + 4BS^2h\right]/D$ \\[6pt]
TP+SP & $\left[\left(4\!\left(1+\tfrac{1}{g}\right)+6F\right)BSh^2 + 4BS^2h\right]/(T\Sigma)$ \\
\bottomrule
\end{tabularx}
\end{table}

\subsection{Practical trade-offs}
TP is attractive when weight-proportional memory (parameters, gradients, optimizer states) is the main bottleneck and high-bandwidth all-reduces are cheap, but it does not reduce activation memory and is therefore ineffective at long sequence lengths. SP is attractive for long-context workloads because it reduces activation memory and keeps the MLP local, but it fully replicates weights and still requires gradient synchronization. TP+SP offers independent control over the two sharding axes but inherits both TP-style reductions and SP-style K/V exchange. TSP is the only scheme that reduces both weight-proportional and activation memory by $1/D$ on a single axis without requiring $D^2$ total devices. Its downside is runtime weight movement, so it is most attractive when memory replication is the dominant bottleneck and weight traffic can be overlapped with compute.

\subsection{Regime analysis}
The key trade-off for TSP is that it eliminates both weight replication and sequence replication, but introduces a weight-movement term per layer. In the large-$D$ regime where $\tfrac{D-1}{D}\approx 1$, comparing TSP and TP forward communication reduces to
\begin{equation}
P_{\mathrm{attn}}\beta_P + P_{\mathrm{mlp}}\beta_P + \frac{2BSh\beta_P}{g}
\;<\;
4BSh\beta_P
\end{equation}
whenever
\begin{equation}
P_L
\;<\;
2BSh\!\left(2-\frac{1}{g}\right).
\end{equation}
For MHA ($g=1$), this simplifies to
\begin{equation}
P_L < 2BSh
\quad\Longleftrightarrow\quad
(4+3F)h < 2BS.
\end{equation}
With $F=4$, this becomes $16h < 2BS$, or equivalently $BS > 8h$. Thus, at long context lengths or moderate batch sizes, we expect TSP to be communication-competitive with TP despite moving weights at runtime.

\subsection{TSP versus SP}

In the forward pass, TSP and SP share the same activation-communication term,
$2BSh\beta_P\,\tfrac{D-1}{D}/g$. TSP additionally moves weights at runtime,
with per-layer cost
$P_{\mathrm{attn}}\beta_P + P_{\mathrm{mlp}}\beta_P\,\tfrac{D-1}{D}$,
whereas SP leaves weights replicated and therefore has no weight-movement term.

During training, both schemes require gradient communication, but for different
reasons. In SP, the full model is replicated across the sequence-parallel group,
so parameter gradients must be synchronized by an all-reduce of cost
$2P_L\beta_g\,\tfrac{D-1}{D}$. In TSP, the parameters are sharded, but each shard
is broadcast to, or rotated through, all sequence shards during the forward pass.
Thus backward produces partial gradients for each weight shard on multiple ranks,
which must be reduced back to the owning rank. This operation is naturally a
reduce-scatter or sum-to-owner reduction rather than an all-reduce, with tight
cost $P_L\beta_g\,\tfrac{D-1}{D}$ (half of the all-reduce volume); for direct
comparison with SP, we conservatively upper-bound its cost by
$2P_L\beta_g\,\tfrac{D-1}{D}$.

Under this accounting, the training communication costs are
\begin{align}
C_{\mathrm{TSP}}^{\mathrm{fwd+bwd}}
&\;\lesssim\;
2P_{\mathrm{attn}}\beta_P
+ 2P_{\mathrm{mlp}}\beta_P\,\tfrac{D-1}{D} \notag \\
&\quad + \frac{4BSh\beta_P}{g}\,\tfrac{D-1}{D}
+ 2P_L\beta_g\,\tfrac{D-1}{D},
\end{align}
\begin{equation}
C_{\mathrm{SP}}^{\mathrm{fwd+bwd}}
=
\frac{4BSh\beta_P}{g}\,\tfrac{D-1}{D}
+ 2P_L\beta_g\,\tfrac{D-1}{D}.
\end{equation}
Thus, when $\beta_P \approx \beta_g$ and $D$ is large, TSP and SP have comparable
training communication up to the additional weight-movement terms in TSP. The
principal advantage of TSP over SP is therefore not lower communication volume
but lower memory: TSP reduces parameter, gradient, and optimizer-state memory by
$1/D$, whereas SP leaves all weight-proportional memory fully replicated.

\subsection{Recomputation communication overhead}
When activation recomputation is employed, the additional forward-pass communication during the backward pass can be significant. For TP, full recomputation increases the total training communication by 50\% (from $8BSh\beta_P\,\tfrac{D-1}{D}$ to $12BSh\beta_P\,\tfrac{D-1}{D}$). For SP and TSP, the overhead is proportionally similar but includes only the attention K/V exchange component, since the MLP is local in SP and the MLP ring in TSP can be overlapped. In all cases, selective recomputation incurs less overhead than full recomputation, since only the attention sub-layer communication is repeated.

\subsection{Communication and computation overlap}

The practical viability of TSP depends not only on total communication volume but on how much of that communication can be hidden behind computation. Our implementation uses explicit overlap strategies for both sublayers.
 
For the gated MLP, the ring schedule is naturally amenable to overlap similar to schemes such as \cite{liu2023ringattention}. At step $t$, each rank launches asynchronous point-to-point send/recv operations for the next weight shards on a dedicated communication stream while computing the current local projections and gated accumulation on the default compute stream. The rank waits on communication only when the next shard is needed, so weight communication is pipelined with the dominant GEMM compute.
 
For attention, each forward pass consists of three stages: broadcasting one rank's packed parameter shard, computing local $Q,K,V$ projections, and all-gathering the corresponding $K,V$ activations before FlashAttention. We pipeline these stages explicitly using one communication stream for weight broadcasts and a second for $K,V$ all-gathers, while the default stream performs projections, FlashAttention kernels, and output GEMMs. The schedule is staggered so that at step $r$, FlashAttention for shard $r\!-\!1$ overlaps with the all-gather for shard $r$ and the broadcast for shard $r\!+\!1$.

\section{Cluster Setup}
\label{sec:cluster-setup}
\vspace{1ex}

\begin{figure*}[htbp]
    \centering
    \begin{minipage}[b]{0.48\textwidth}
        \centering
        \includegraphics[width=\textwidth]{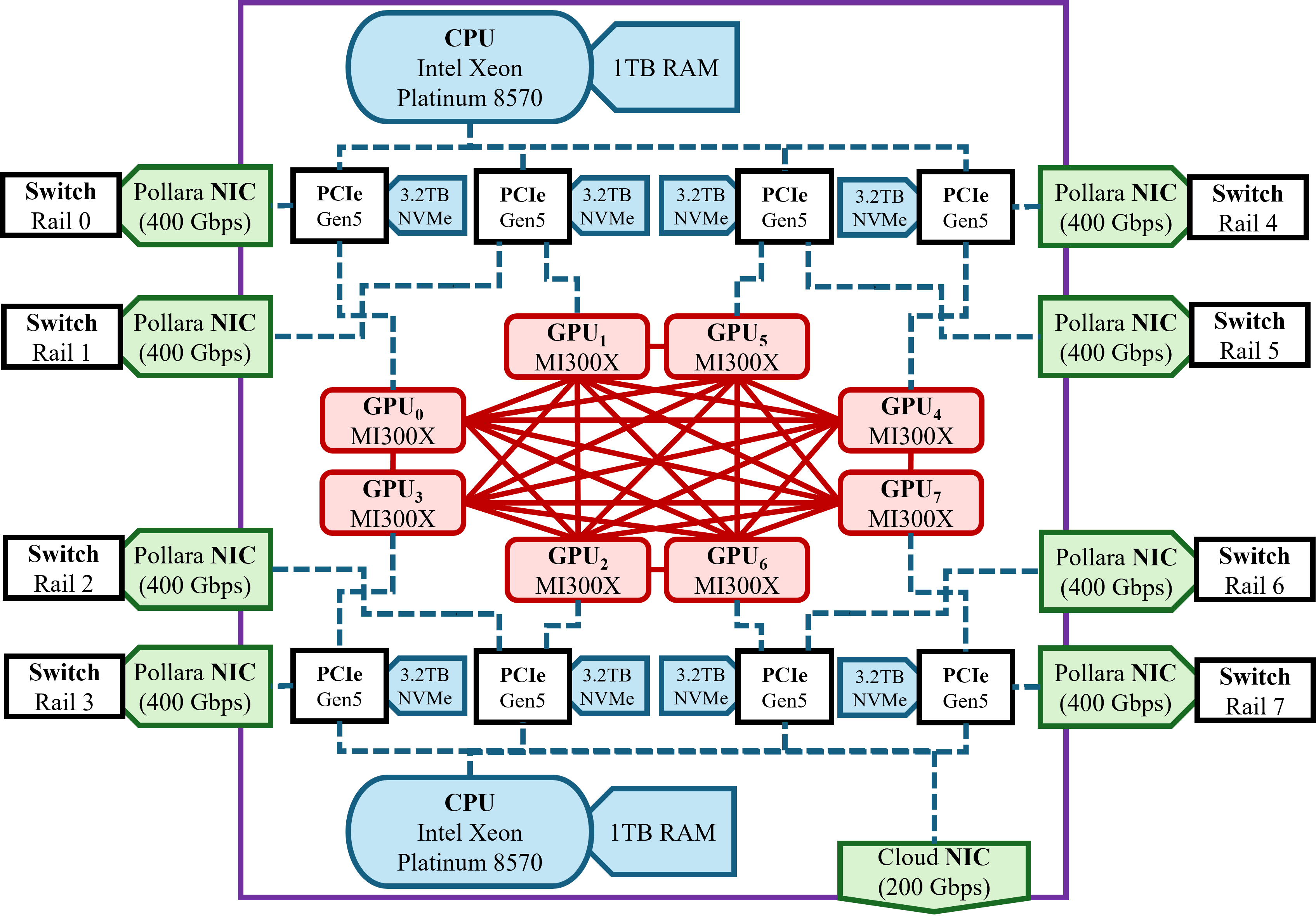}
        \caption*{(a) The architecture of a single node}
    \end{minipage}
    \hfill
    \begin{minipage}[b]{0.50\textwidth}
        \centering
        \includegraphics[width=\textwidth]{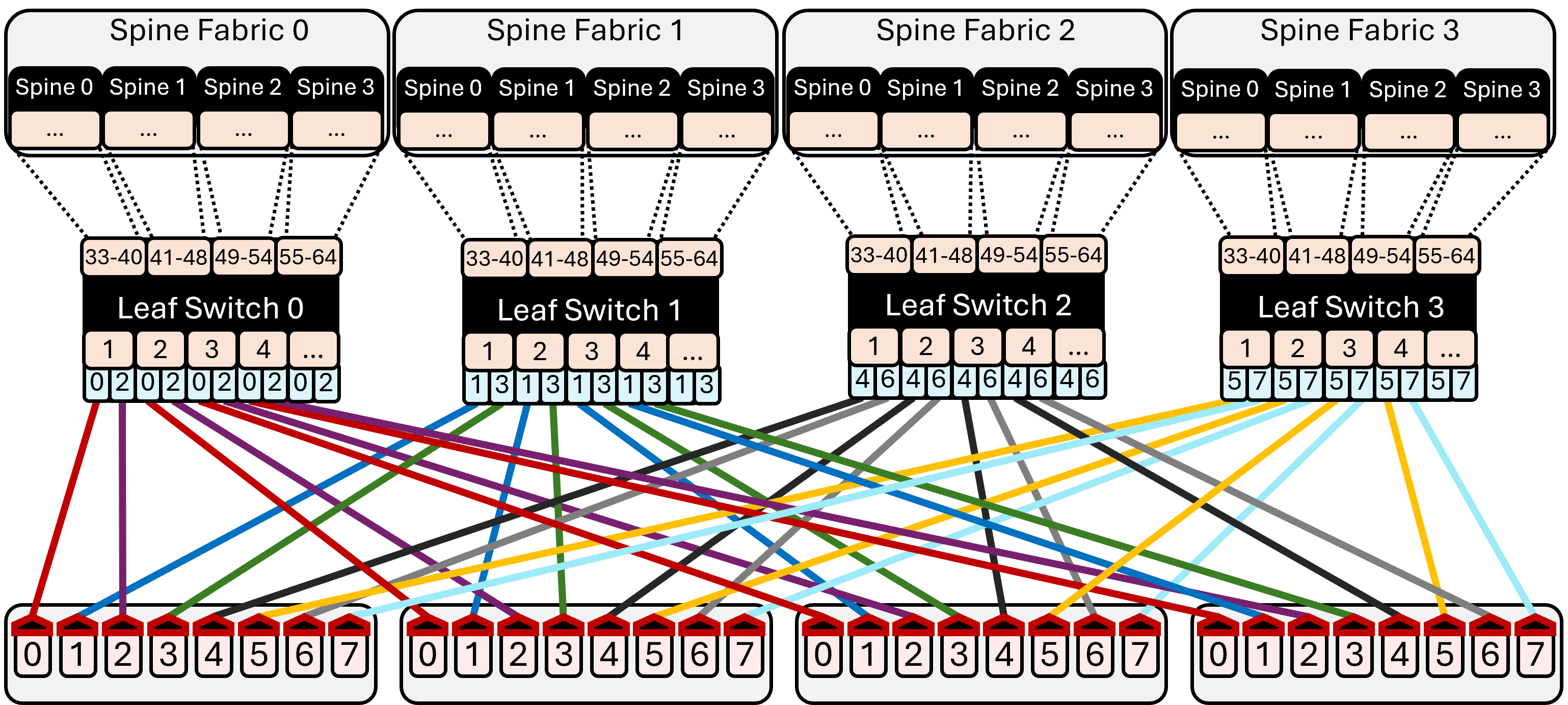}
        \caption*{(b) Cluster topology}
    \end{minipage}
    \caption{The architecture of the Zyphra pretraining cluster. Each node contains 8 MI300X GPUs interconnected with Infinity Fabric. Each GPU is assigned its own Pollara 400 Gbps NIC, which connects the GPU to its respective switch in a rails-only topology.  Our cluster switch topology is a two-level rails-only setup, with four physical leaf switches, and four spine switches per leaf switch. We have 4 physical leaf switches per slice. Each slice consists of 30 servers. The slices are aggregated via 4 spine switches per leaf. Each compute node is connected to the head-node and storage node via a separate interconnect to avoid contention of data/checkpoint loading and training comms.}
    \label{fig:cluster-arch}
\end{figure*}

Tensor parallelism (TP) and sequence parallelism (SP) are ultimately hardware-aware schemes, and our analysis of any parallelism requires a discussion of how it maps to available hardware. To facilitate such discussions, we describe our cluster in this section as well as Appendix \ref{app:cluster}.

\subsection{Node Architecture}
\label{sec:node-architecture}

Each compute node consists of eight MI300X GPUs connected via Infinity Fabric, dual-socket Intel Xeon CPUs with 2\,TB of DDR5 memory, a dedicated Pollara 400~Gbps NIC per GPU, and local NVMe storage for high-throughput datasets and checkpoints (see Figure~\ref{fig:cluster-arch}). A detailed description for the compute, storage, and login nodes is provided in Table~\ref{tab:hardware_specs} and Appendix~\ref{app:cluster}.

\begin{table*}[ht]
\centering
\caption{Hardware specifications for compute, storage, and login nodes. Each node type has separate local drives for the OS.}
\label{tab:hardware_specs}
\begin{tabular}{|c|c|c|c|}
\hline
\textbf{Hardware Element} & \textbf{Compute Node} & \textbf{Storage Node} & \textbf{Login Node} \\
\hline
\textbf{GPUs} & 8 MI300X & --- & --- \\
& \citep{amd2024cdna3} & & \\
\hline
\textbf{RAM} & 2\,TB DDR5 & 256\,GB DDR5 & 80\,GB \\
& 16$\times$128\,GB DIMMs & 16$\times$16\,GB DIMMs & 5$\times$16\,GB DIMMs \\
& Samsung M321RAJA0MB0-CWMNY & Samsung M321R2GA3BB6-CQKET & (virtual/QEMU) \\
& 5600 MT/s & 4800 MT/s & \\
\hline
\textbf{CPU} & 2$\times$ Intel Xeon Platinum 8570 & 2$\times$ Intel Xeon Gold 6426Y & 2$\times$ Intel Xeon \\
& 56 cores, 2 threads/core & 16 cores, 2 threads/core & (Sapphire Rapids) \\
& 1\,TB RAM per socket & 128\,GB RAM per socket & 8 cores, 2 threads/core \\
\hline
\textbf{Networking} & 8$\times$ Pollara 400 NICs (400Gbps) & 1$\times$ Pensando DSC NIC (200Gbps) & --- \\
\hline
\textbf{Storage} & 25.6\,TB & 120\,TB & \\
& 8$\times$ NVMe drives (3.2\,TB each) & 16$\times$ NVMe drives ($\approx$7.6\,TB each) & $\approx$1\,TB \\
& Micron MTFDKCC3T2TGQ & RAID0 array (\texttt{/dev/md127}) & \\
\hline
\end{tabular}
\end{table*}

\subsection{Cluster Architecture}

Instead of the classical Clos \citep{clos} topology, individual nodes are interconnected in a rails-only topology, which was popularized in \citet{meta-rails} as a way to exploit the nature of 3D-parallel deep learning training communication patterns to reduce the number of Arista 7060X6-64PE \citep{arista7060x6} switches. Specifically, our cluster switch topology is a two-level rails-only setup, with four physical leaf switches, and four spine switches per leaf switch. This design trades some path diversity and routing flexibility compared to a full Clos fabric for lower cost on physical switches and simpler wiring. In practice, this means the topology leads to a slight performance penalty compared to a Clos network unless the parallelism topology and collective communication algorithm is carefully designed to avoid cross-rail traffic.

Our cluster employs two physically separate networks. The Pollara training fabric handles all collective communication for distributed training, while a VPC network manages dataset I/O, checkpoints, and cluster management. This separation prevents storage operations from interfering with model communication (gradients, optimizer states, etc). The VPC network connects each node via Pensando DSC SmartNIC/DPUs to a separate leaf--spine fabric, providing approximately 200 Gbps throughput to the storage node from any of the compute nodes.

\section{Results}
\label{sec:results}

We now evaluate the empirical behavior of TSP using the implementation described in \S\ref{sec:design}. All experiments are conducted on the cluster described in \S\ref{sec:cluster-setup} using MI300X GPUs with Infinity Fabric intra-node interconnect and Pollara inter-node interconnect.
 
\subsection{Peak memory versus sequence length}
 
\begin{figure}[t]
    \centering
    \includegraphics[width=0.99\columnwidth]{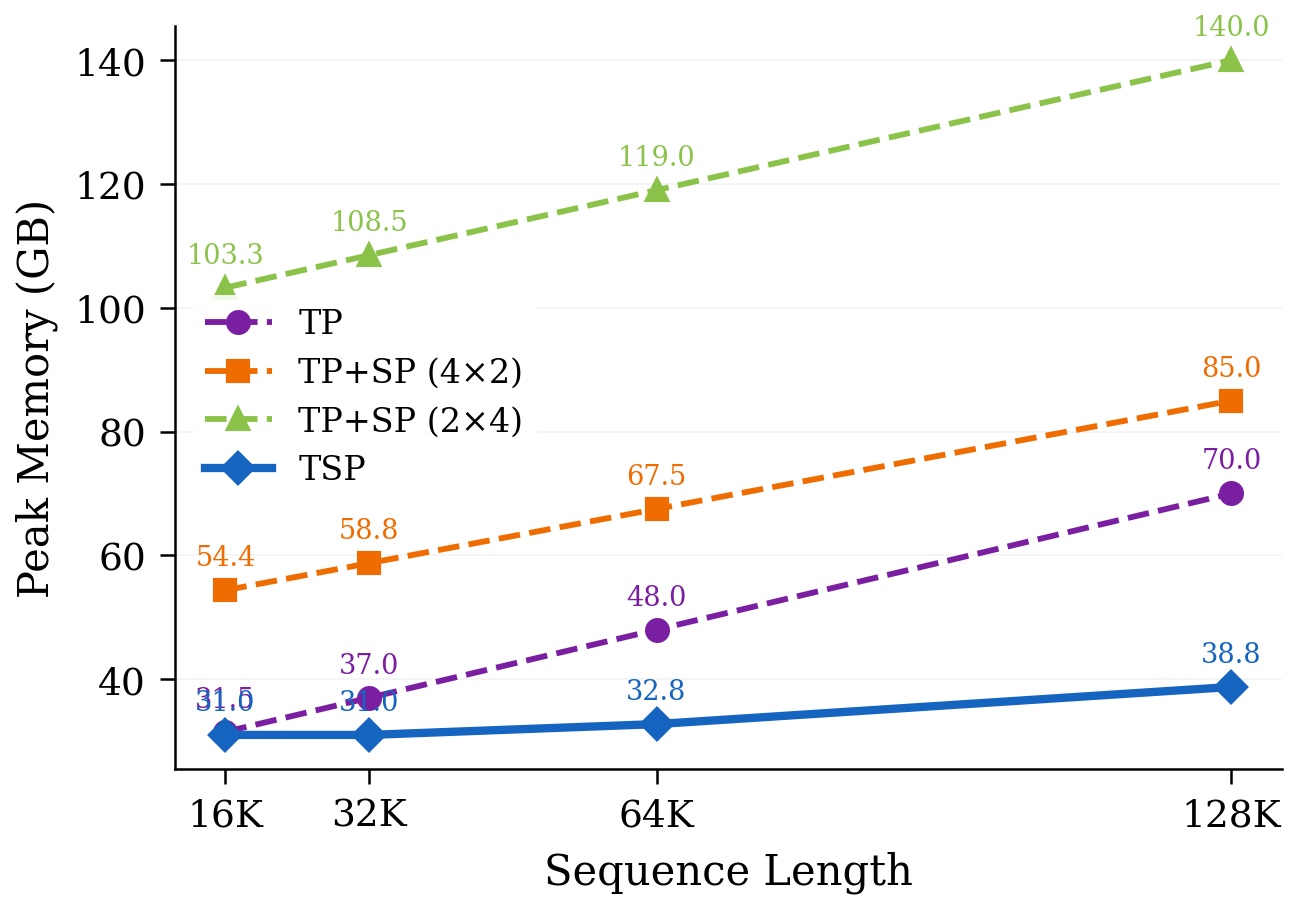}
    \caption{Per-GPU peak memory as a function of sequence length for TP, TP+SP (two factorizations), and TSP, all on a single node (8 MI300X GPUs). TSP provides the lowest peak memory across all tested sequence lengths, with the advantage widening as context grows.}
    \label{fig:peak-mem-vs-seqlen}
\end{figure}
 
Figure~\ref{fig:peak-mem-vs-seqlen} shows per-device peak memory across sequence lengths from 16K to 128K tokens. TSP yields the lowest peak memory among all measured strategies at every point. At 16K tokens, TSP and TP are nearly identical (31.0\,GB versus 31.5\,GB), consistent with the short-context regime in which weight-proportional memory dominates. We expect TSP to have the same weight sharding behavior as TP (which is why the initial point is nearly even with TP), and for the TSP sequence length slope to match that of SP.
 
This behavior matches the memory model in \S\ref{sec:comm-mem-model}. TP reduces only weight-proportional terms within TP-supported blocks, SP reduces only the activation term, and TSP applies a single $1/D$ sharding factor to both terms simultaneously. The results also show that for TP+SP layouts, memory is determined by the split between TP and SP. Specifically, when context is short, a larger TP degree performs better because model-state memory dominates. However at long context, the activation term dominates and the advantage shifts only partially toward larger SP degree.
 
\subsection{Throughput versus sequence length}
 
\begin{figure*}[t]
    \centering
    \includegraphics[width=.75\textwidth]{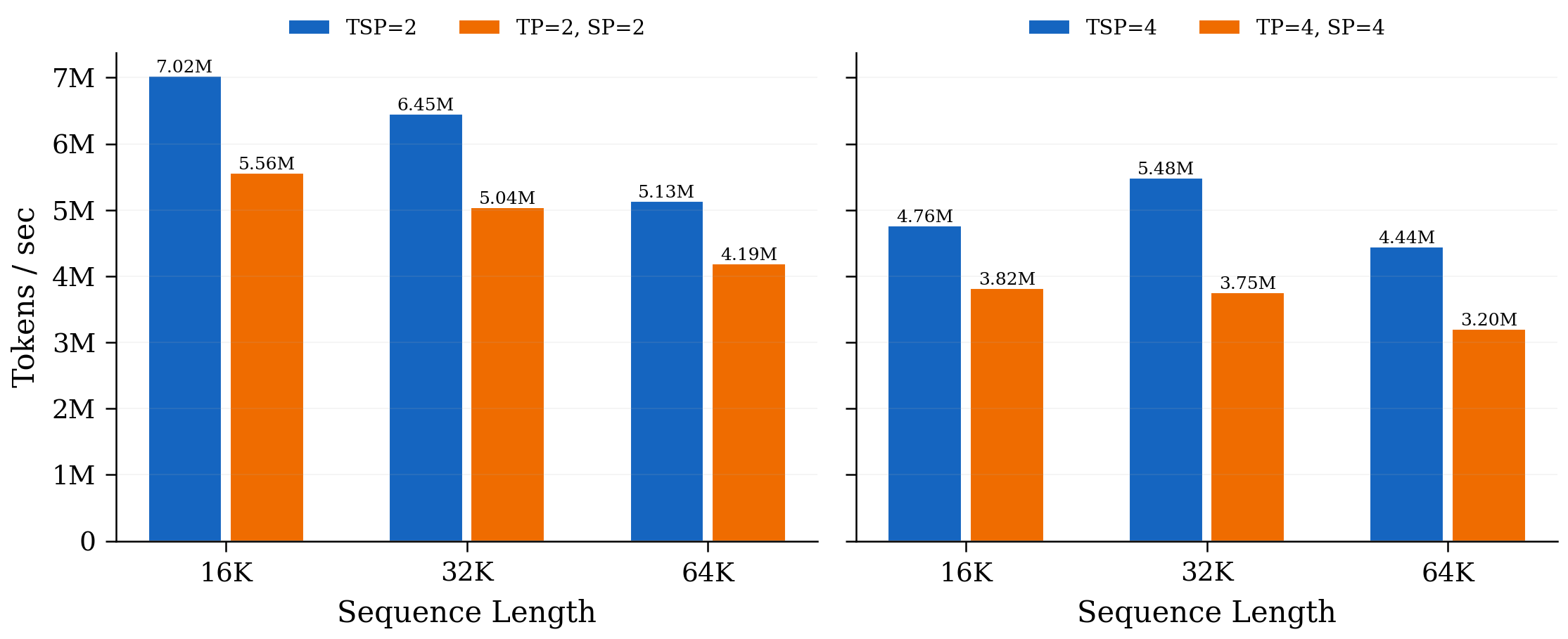}
    \caption{Forward-pass throughput (tokens/s) versus sequence length for TSP and matched TP+SP baselines at folded degrees $D{=}2$ (left) and $D{=}4$ (right), on 2 full nodes (16 MI300X GPUs).}
    \label{fig:tput-vs-seqlen-short}
\end{figure*}
 
\begin{figure*}[t]
    \centering
    \includegraphics[width=.95\textwidth]{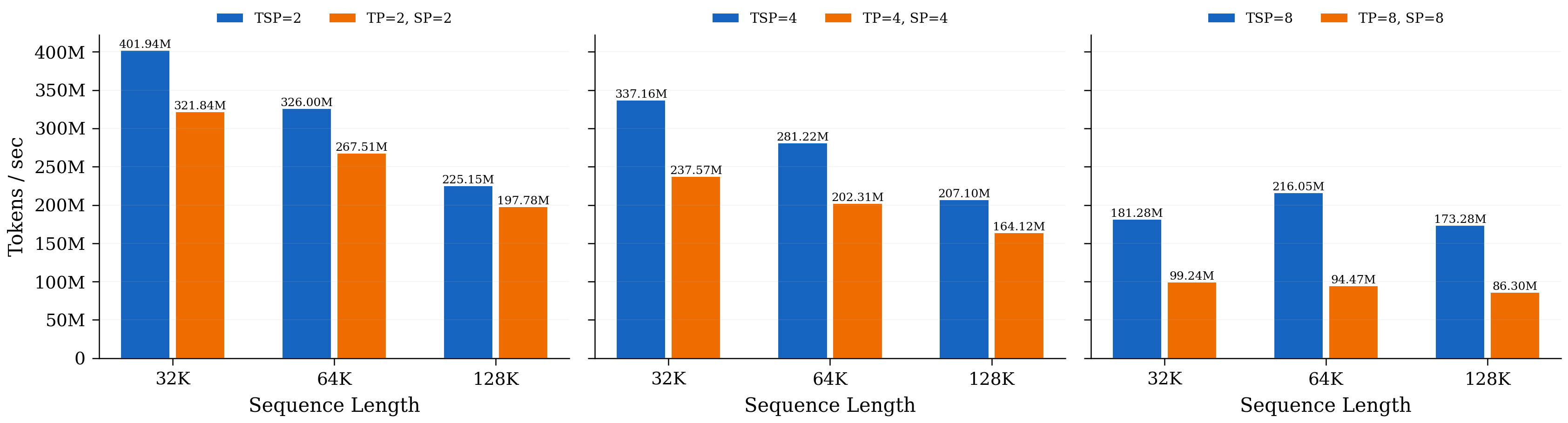}
    \caption{Forward-pass throughput (tokens/s) versus sequence length at degrees $D{=}2$, $D{=}4$, and $D{=}8$ over the 32K--128K range, on 128 full nodes (1024 MI300X GPUs). The TSP advantage grows with increasing degree.}
    \label{fig:tput-vs-seqlen-long}
\end{figure*}
 
We collect total throughput measurements of the forward pass in Figures~\ref{fig:tput-vs-seqlen-short} and~\ref{fig:tput-vs-seqlen-long}. In the sequence-length sweeps at $D{=}2$ and $D{=}4$ (Figure~\ref{fig:tput-vs-seqlen-short}), TSP consistently outperforms the matched TP+SP baseline. A similar trend appears at longer context in Figure~\ref{fig:tput-vs-seqlen-long}. Increasing the folded degree does not collapse TSP throughput. These empirical measurements indicate that the theoretical extra communication volume traffic does not lead to a significant increase in wall-clock time. Weight transfers are pipelined behind the dominant GEMMs using the overlap strategy described in \S\ref{sec:comm-mem-model}, so they consume bandwidth without consuming critical-path time.
 
\subsection{Throughput versus micro-batch size}
 
\begin{figure*}[t]
    \centering
    \includegraphics[width=\textwidth]{figures/fig_train_2node.png}
    \caption{Forward+backward throughput (tokens/s) versus micro-batch size at degrees $D{=}2$, $D{=}4$, and $D{=}8$ on 2 full nodes (16 MI300X GPUs). TSP maintains a consistent advantage over TP+SP across batch sizes.}
    \label{fig:tput-vs-mbs}
\end{figure*}
 
Figure~\ref{fig:tput-vs-mbs} presents a micro-batch-size sweep at fixed sequence length. Because the folded layout reduces both model-state memory and activation memory, larger micro-batches can fit into device memory, which improves throughput and reduces per-replica parallelism.

\section{Discussion}
\label{sec:discussion}

The principal benefit of TSP is reduced replication of both weights and activations. The results above suggest that in the long-context and memory-constrained settings that motivated this work, eliminating replication matters more than minimizing communication volume.

Communication volume and communication cost are not the same thing: The theoretical analysis in \S\ref{sec:comm-mem-model} predicts higher volume for TSP, but how much that translates into wall-clock cost depends on whether collectives are latency- or bandwidth-bound, how much traffic can be overlapped with matrix multiplication, and how much memory headroom remains for larger problem sizes such as increased batch size.
 
In the peak memory results, TSP remains close to TP in the short-context regime, where weight-proportional memory dominates, and then increasingly diverges in the long-context regime as activations become dominant. TSP inherits TP's reduced memory at short context and SP's reduced memory at long context, without requiring the two-dimensional device allocation of conventional TP+SP.
 
Because folding reduces the number of devices in the combined model-parallel group, the entire folded model replica can more easily fit within the high-bandwidth intra-node domain. This is beneficial in some combinations of models and sequence lengths despite incurring more raw communication volume, since the alternative in a conventional layout TP, SP, or TP+SP may require one of the parallelism axes to span the slower inter-node links. The crossover point between TSP and conventional parallelism schemes will depend on the ratio of accelerator FLOPs to intra- and inter-node communication bandwidth, and our results are obtained on hardware where intra-node bandwidth is especially favorable.

TSP composes orthogonally with the existing axes of multi-dimensional parallelism. Because the TSP group is a single mesh dimension, it can be combined with additional TP, SP, PP, EP, or DP axes and existing mesh implementations. As a concrete example, suppose a workload calls for a tensor-sharding factor of $8$ and a sequence-sharding factor of $2$. A conventional layout uses $(\mathrm{TP}{=}8) \times (\mathrm{SP}{=}2) = 16$ model-parallel ranks; if a node holds $8$ GPUs, this forces one axis across inter-node links. The same effective sharding can instead be expressed as $(\mathrm{TP}{=}4) \times (\mathrm{TSP}{=}2) = 8$ ranks, fitting entirely within a node and freeing the remaining $8$ ranks to expand the data-parallel mesh. More generally, TSP can be inserted as a flexible axis wherever the parallelism budget would otherwise force model-parallel groups across slow links, which makes it a useful addition to the pool of potential parallelism schemes and enables easier optimization to future or unforeseen hardware topologies.

\section{Related Work}
\label{sec:related-work}

\paragraph{Parallelism}
Tensor parallelism (TP) for Transformers was established in Megatron-LM~\citep{shoeybi2019megatron}. TP has since become an important axis of scale-up model parallelism in works such as ~\cite{narayanan2021efficient}. Sequence-parallel (SP) variants further shard activations along the sequence dimension to reduce memory~\citep{korthikanti2023reducing}, while Mesh-TensorFlow formalized the assignment of parallelism dimensions to orthogonal mesh axes~\citep{shazeer2018meshtensorflow}. Long-context methods extend this line in three main directions: ring-based attention, which circulates KV blocks point-to-point and overlaps communication with blockwise attention~\citep{liu2023ringattention,li2021sequenceparallelism}; all-gather-based sequence parallelism, which exchanges all KV shards in a single collective before running attention against the full-sequence keys and values; and all-to-all-based context parallelism, which redistributes QKV so each device attends over the full sequence for a subset of heads~\citep{jacobs2023deepspeedulysses}. Subsequent schemes combine these ideas with round-robin or zigzag token layouts to better load-balance causal-attention ~\citep{brandon2023stripedattention,grattafiori2024llama3,fang2024unifiedsp}.

\paragraph{Folding and MLP Ring communication}
The most direct inspiration for our work is MoE Parallel Folding by~\cite{liu2026moeparallelfolding}, which decouples the parallel mappings used by attention and MoE blocks and folds communication-intensive dimensions into compact groups, such that different block types can adopt different parallel layouts. TSP adopts the folding intuition in the dense transformer setting for TP and SP specifically. Instead of placing TP and SP on separate mesh axes, TSP collapses them onto a single folded dimension. This also distinguishes TSP from prior ring-attention methods, which circulate activations while keeping MLP computation local with replicated weights. The closest prior work in that direction is METP by~\cite{liang2025metp}, which applies ring-based point-to-point communication to both attention and MLP layers for long-sequence training.

\section{Conclusion}
\label{sec:conclusion}

We have presented tensor and sequence parallelism (TSP), a parallel execution strategy that folds tensor parallelism and sequence parallelism onto a single device axis. By assigning each rank both a weight shard and a sequence shard, TSP reduces both per-device parameter-proportional (parameters, optimizer states, gradients) memory and activation memory. Our theoretical analysis characterizes the expected communication and memory trade-offs, and we have validated these with an empirical evaluation on up to 1024 MI300X GPUs. TSP achieves the lowest peak memory compared to TP, SP, and TP+SP meshes across all tested sequence lengths while maintaining competitive throughput. The advantage widens with increasing parallelism degree and sequence length. These results position TSP as a practical, topology-aware alternative for long-context and memory-constrained model training.

\clearpage

\clearpage

\bibliographystyle{tmlr}
\bibliography{main}

\clearpage

\appendices

\section{Cluster Details}
\label{app:cluster}

\subsection{Compute Nodes}

Each compute node contains:

\begin{itemize}
    \item \textbf{GPUs}: 8 MI300X~\citep{amd2024cdna3} GPUs connected via Infinity Fabric intra-node interconnect
    \item \textbf{RAM}: 2\,TB of DDR5 RAM. Specifically, 16x128\,GB DIMMs of Samsung M321RAJA0MB0-CWMNY, running at 5600 MT/s.
    \item \textbf{CPU}: 2 physical sockets of Intel Xeon Platinum 8570, each with 56 physical cores and 2 threads per core. Each socket is connected to 1\,TB of RAM (8 DIMMs)
    \item \textbf{Networking Cards}: Eight Pollara 400~\citep{amd_pollara_2024} network interface cards (NICs), each at 400Gbps. One Pensando DSC 200 GbE cloud NIC for loading data and checkpoints. 
    \item \textbf{Storage}: 25.6\,TB split into 8 physical NVMe drives (Micron MTFDKCC3T2TGQ-1BK1DABDB), each with 3.2\,TB.
\end{itemize}

\subsubsection{Storage Node}

The storage node contains:

\begin{itemize}
    \item \textbf{RAM}: 256\,GB of DDR5 RAM. Specifically, 16x16\,GB DIMMs of Samsung M321R2GA3BB6-CQKET, running at 4800 MT/s.
    \item \textbf{CPU}: 2 physical sockets of Intel(R) Xeon(R) Gold 6426Y, each with 16 physical cores and 2 threads per core. Each socket is connected to 128\,GB of RAM (8 DIMMs).
    \item \textbf{Networking Cards}: One Pensando DSC 100 GbE cloud NIC for loading data and checkpoints.
    \item \textbf{Storage}: 120\,TB configured as a single RAID0 array (`/dev/md127`) from 16 physical NVMe drives \\ (Micron\_7450\_MTFDKCC7T6TFR), each with ~7.6\,TB.
\end{itemize}

Each of the above nodes have separate local drives for the OS.

\subsubsection{Login Node}

The login node is a VM that contains:

\begin{itemize}
\item \textbf{RAM}: 80\,GB system memory, installed as 5×16\,GB DIMMs (virtual/QEMU).
\item \textbf{CPU}: 2 sockets of Intel Xeon (Sapphire Rapids), each with 8 cores and 2 threads per core (virtualized under KVM).
\item \textbf{Storage}: ~1\,TB total, split into 3 virtual disks: vda (100\,GB), vdb (520\,GB), and vdc (520\,GB)
\end{itemize}

\section{Author Contributions}
\label{app:contributions}

\paragraph{Vasu Shyam} Conceptualization; algorithm design and development; initial correctness verification.
\paragraph{Anna Golubeva} Large-scale experimental evaluation; collection and organization of experimental data.
\paragraph{Quentin Anthony} Theoretical analysis and cost modeling; manuscript writing and preparation; final figure and plot formatting.

\section{Reference Model Configuration}
\label{app:model-config}

Unless otherwise stated, all theoretical figures and analyses in this paper use the following 7B dense decoder-only transformer configuration.

\begin{table}[h]
\centering
\begin{tabular}{lc}
\toprule
\textbf{Hyperparameter} & \textbf{Value} \\
\midrule
Hidden / residual dimension $h$ & 4096 \\
Number of layers $L$ & 32 \\
Number of query heads $n_h$ & 32 \\
Number of K/V heads $n_{kv}$ & 32 (MHA, $g=1$) \\
Per-head dimension $d$ & 128 \\
FFN expansion factor $F$ & 4 (gated MLP) \\
Forward precision $\beta_P$ & 2 bytes (bf16/fp16) \\
Gradient precision $\beta_g$ & 2 bytes \\
Optimizer states $n_o$, $\beta_o$ & 3, 4 bytes (mixed-precision AdamW) \\
\bottomrule
\end{tabular}
\end{table}

The total parameter count under this configuration is $L \cdot (4 + 3F) h^2 \approx 8.6$~B, ignoring embedding and norm parameters; we refer to it as the "7B reference model" in the main text.

\end{document}